\pdfoutput=1

\documentclass[11pt]{article}

\usepackage{acl}

\usepackage{times}
\usepackage{latexsym}

\usepackage[T1]{fontenc}

\usepackage[utf8]{inputenc}

\usepackage{microtype}

\usepackage{adjustbox}
\usepackage{booktabs}
\usepackage{tabularx}
\usepackage{multirow}
\usepackage{siunitx}
\usepackage{rotating}
\usepackage[disable]{todonotes}
\usepackage{makecell}
\usepackage{amsmath}
\usepackage{subcaption}
\usepackage{xspace}
\usepackage{xcolor}
\usepackage{framed}
\usepackage{soul}
\usepackage{diagbox}
\usepackage{makecell}

\usepackage{acro}
\usepackage{graphicx}
\usepackage{enumitem}
\usepackage{footmisc}
\usepackage[normalem]{ulem}
\usepackage{amssymb}

\newcolumntype{L}[1]{>{\raggedright\let\newline\\\arraybackslash\hspace{0pt}}m{#1}}
\newcolumntype{C}[1]{>{\centering\let\newline\\\arraybackslash\hspace{0pt}}m{#1}}
\newcolumntype{R}[1]{>{\raggedleft\let\newline\\\arraybackslash\hspace{0pt}}m{#1}}

\DeclareAcronym{am}{
	short = TDAM,
	long = Topic-Dependent Argument Mining
}

\DeclareAcronym{ukpc}{
	short = UKP Corpus,
	long = UKP Sentential Argument Mining Corpus
}
\DeclareAcronym{fsc}{
	short = FS150T-Corpus,
	long = Few-Shot-150T Corpus
}
\DeclareAcronym{iam}{
	short = IAM-Corpus,
	long = IAM-Corpus
}
\DeclareAcronym{ibm}{
	short = IBM-Corpus,
	long = IBM-Corpus
}

\title{Diversity Over Size: On the Effect of Sample and Topic Sizes for Topic-Dependent Argument Mining Datasets}

\author{Benjamin Schiller$^{a,b}$, Johannes Daxenberger$^a$, Andreas Waldis$^{b, c}$ \and Iryna Gurevych$^b$ \\
    $^a$summetix GmbH, $^b$Ubiquitous Knowledge Processing Lab,\\ Department of Computer Science, Technical University of Darmstadt, \and \\ $^c$Information Systems Research Lab, Lucerne University of Applied Sciences and Arts \\ $^a$\texttt{\{schiller, daxenberger\}@summetix.com},\\ $^b$\url{www.ukp.tu-darmstadt.de},\\$^c$\url{www.hslu.ch}}

\begin{document}
\maketitle
\begin{abstract}
\ac{am}, that is extracting and classifying argument components for a specific topic from large document sources, is an inherently difficult task for machine learning models and humans alike, as large \ac{am} datasets are rare and recognition of argument components requires expert knowledge. 
The task becomes even more difficult if it also involves stance detection of retrieved arguments.
In this work, we investigate the effect of \ac{am} dataset composition in few- and zero-shot settings.
Our findings show that, while fine-tuning is mandatory to achieve acceptable model performance, using carefully composed training samples and reducing the training sample size by up to almost 90\% can still yield 95\% of the maximum performance. 
This gain is consistent across three \ac{am} tasks on three different datasets.
We also publish a new dataset\footnote{\url{https://tudatalib.ulb.tu-darmstadt.de/handle/tudatalib/4353}} and code\footnote{\url{https://github.com/UKPLab/argument-topic-diversity}} for future benchmarking.
\end{abstract}

\section{Introduction}
\acf{am} is the task of extracting argument components in documents or document collections \cite{DBLP:journals/tacl/LauscherWGG22}. 
Topic-dependence (or, as \citet{Stab2018b} refer to it, \emph{information-seeking} argument mining) means that argument components are directed towards a given topic \cite{ibm-corpuswide,shnarch-etal-2018-will}.
The topic is used in two ways: by a machine learning model to learn topic-relevance and as a query to retrieve input documents for automatic argument search \cite{mci/Daxenberger2020}.

While LLMs (large language models) show astounding results \cite{touvron2023llama,openai2023gpt4}, task-related datasets are still important to improve model performance \cite{dettmers2023qlora,LOMO2023parameter,NEURIPS2022_0cde695b,van-der-meer-etal-2022-will} and decrease certain undesirable behaviours \cite{NEURIPS2022_b1efde53,askell2021general} via fine-tuning, 
and to provide curated data for evaluation purposes.
To assemble large amounts of training samples, it is common to use non-experts to annotate datasets.
However, in contrast to a task like sentiment analysis, the task of identifying arguments is not naturally understood by non-experts and, due to pitfalls like commonly used fallacies \cite{habernal-etal-2018-fallacies}, needs a thorough training phase and strict quality control of the crowdsourcing process. Hence, crowdsourcing datasets for \ac{am} is not only time-consuming but also expensive, as it requires a large number of workers per sample for satisfactory agreement. For instance, \citet{Stab2018b} report a sum of \$2,774 for the annotation of 25,492 samples, requiring seven annotators to reach a satisfying inter-annotator agreement.

Due to the efficacy of transformers \cite{vaswani2017attention}, datasets for \ac{am} (as for many other tasks) have grown in size over recent years \cite{Stab2018b,shnarch-etal-2018-will,rinott-etal-2015-show,aharoni-etal-2014-benchmark}. 
Recent datasets for \ac{am} contain up to 30,000 samples \cite{ibm-corpuswide}.  
However, relying on large datasets has several disadvantages: (1) it is impractical to label such large datasets by experts, (2) crowdsourcing them is costly, and (3) training (as well as tuning) takes longer and adds to the cost. 

To tackle those disadvantages, we study if and how dataset sizes for \ac{am} can be reduced and what the \textit{composition} (total number of topics, samples, and samples per topic) of these datasets should be to train high-performing models. In contrast to simpler text classification tasks with a single input (e.g. document categorization or sentiment analysis), creating datasets for cross-topic \ac{am} is more complicated, as it requires controlling two or more inputs (e.g. topic and argument component) and a diverse choice of topics, which we show in this work.
\\\\
Our work is motivated by few-shot learning \cite{wei2021finetuned,schick-2021-cloze,ruckle-emnlp-2020-zeroshot,NIPS2016_90e13578} and diversity sampling \cite{larson-etal-2019-outlier,katharopoulos2018not,NIPS2017_2f37d101} approaches. \citet{larson-etal-2019-outlier} show that \textit{unique} samples (similar to \textit{outliers} in \citet{NIPS2017_2f37d101}), i.e. samples that differ strongly in structure or content from other samples, can increase model robustness.
Thus, in addition to relying on models that are able to learn with fewer samples, we increase the diversity of samples in our dataset by integrating a large number of distinct topics (i.e., outliers) and, in turn, aim to increase the robustness of our models.

As a testbed, we create a benchmark with two datasets that have an equal number of training samples and only differ in the number of topics and samples per topic. We research the influence of these two compositional parameters on model performance and costs of the annotation process, showing that a largely increased number of topics improves model performance by up to 4.1pp in this scenario. 
We verify findings from the benchmark datasets on two \ac{am} datasets from different domains with a slightly different task and find that we can save up to almost 90\% of the annotation costs if we are willing to sacrifice 5\% of the maximum model performance.

Our contributions are as follows: (1) We create a new dataset for \ac{am} which differs from an existing \ac{am} dataset, namely the \ac{ukpc} \cite{Stab2018b}, only in the number of topics and samples per topic, allowing for a deeper analysis of this task and assumptions on how future datasets can be composed (diversity sampling), (2) we analyze zero- and few-shot experiments on the new dataset, giving recommendations on efficient dataset composition, (3) we evaluate findings on dataset efficiency on two different \ac{am} tasks from another domain, and (4) we present state-of-the-art results on the \ac{ukpc}.

\section{Related Work}

\textbf{\ac{am}} The task of \textit{Discourse-level} Argument Mining aims to classify argument components \cite{rocha2023crossgenre,ajjour-etal-2017-unit,Stab2014a,goudas2014} and their relations \cite{eger-etal-2017-neural,nguyen-litman-2016-context} within isolated documents. \textit{Topic-Dependent} Argument Mining \cite{DBLP:journals/tacl/LauscherWGG22}, however, describes the task of searching large, heterogeneous document collections for argument components relevant to a given topic \cite{ibm-corpuswide,Stab2018b,shnarch-etal-2018-will}. In this work, we will focus on the latter instead of extracting components like claims and premises or their relations from single documents. 

\textbf{Dataset composition} The growth of sample sizes in \ac{am} datasets seems to be a necessity to cover wider ranges of topics and, thus, to support better cross-topic and cross-domain performance \cite{ibm-corpuswide,Stab2018b,shnarch-etal-2018-will,rinott-etal-2015-show,aharoni-etal-2014-benchmark}. As this, in consequence, increases the annotation and training costs of the models, we aim to uncover low-effort methods for \ac{am} datasets that help to keep the number of training samples as low as possible while reaching similar performance. \citet{ajjour-etal-2023-topic} discover that many \ac{am} datasets, even those with large amounts of samples, do mostly cover topics that frequently appear in forums, but leave out many less-frequently discussed areas. We argue that using too many samples per topic in \ac{am} datasets is a waste of financial resources and focusing on only a few, frequently discussed topics limits the capability of models to generalize well in cross-topic experiments. In this work, we propose a different compositional structure for future \ac{am} datasets. 

\textbf{Model learning techniques} The intuition of keeping the number of training samples---and hence, costs and annotation effort---low, has attracted research that focuses on techniques enabling models to learn with less data. With regard to models, one of the most impactful designs in recent years are transformer \cite{vaswani2017attention} that are pre-trained on large amounts of text with unsupervised learning techniques \cite{robertaliu,devlin2018bert}. These LLMs show remarkable results on few-shot learning \cite{gao2021making,schick-2021-cloze} and zero-shot learning \cite{wei2021finetuned,ruckle-emnlp-2020-zeroshot,radford2019language} tasks. One form of zero-shot learning that gained a lot of attention due to its astounding performance is \textit{prompting} (in-context learning), where a pre-trained model is not fine-tuned and, in addition to the actual input, is only given exemplary inputs (for instance, prepended to the actual input) at inference \cite{gpt3_2020}. Two other and older techniques used to reduce training sample sizes are transfer and multi-task learning, which have also been successfully combined with LLMs \cite{schiller2021stance,liu2019mt-dnn}.

\textbf{Benchmarks and diversity sampling} In contrast to most of these techniques that concentrate on adapting model architectures in a way such that models are able to learn with few or no samples, we focus on benchmarking efficient compositional structures of \ac{am} datasets. In recent years, other benchmarks consisting of multiple datasets have been published \cite{schiller2021stance,wang2019superglue,wang2019glue} with the aim to standardize performance reports for machine learning models and, hence, allowing new model architectures to compete against each other and to make the results comparable. 
\citet{arakelyan-etal-2023-topic} show improvements on a benchmark for the task of stance detection by sampling a subset with a deep, unsupervised topic model and training on the subset with a contrastive objective.
We, however, aim to adapt and benchmark the composition of \ac{am} datasets we train the models with, such that existing models (without any modification) can exploit it. Our decision on how to ensemble the datasets we use in this work draws on insights of diversity sampling research which shows that models can profit from datasets with high diversity, i.e. containing samples that differ strongly from each other \cite{larson-etal-2019-outlier,katharopoulos2018not,NIPS2017_2f37d101}. While other work in the area of \ac{am} has scratched on the topic of diversity by increasing the number of used topics \cite{ibm-corpuswide}, there has been no work that we know of, dedicated on determining the ideal dataset composition for \ac{am} datasets.

\begin{table*}[!hbt]
\scriptsize
\centering
\def\arraystretch{1.5}
\begin{tabularx}{\linewidth}{lcccccL{4.5cm}}
\Xhline{2\arrayrulewidth}
\textbf{Datasets} & \# \textbf{Topics} & \multicolumn{4}{c}{\textbf{\# Samples}} &  \\ \cline{3-6}
 & & \textbf{Train} & \textbf{Dev} & \textbf{Test} & \textbf{Total} & \textbf{Classes}\\ \hline
\acs{fsc} (ours) & 150 & 17,280 & 1,440 & 2,880 & 21,600  & pro (19\%), con (19\%), none (62\%)  \\ 
\acs{ukpc} \cite{Stab2018b} & 8 & 17,280 & 2,475 & 1,249 & 5,481   & pro (19\%), con (24\%), none (56\%) \\
\acs{iam} \cite{cheng-etal-2022-iam} & 100 & 9,678 & 7,057 & 7,065 & 23,800 &  support (11\%), contest (10\%), no relation (79\%) \\
\acs{ibm} \cite{ibm-corpuswide} & 221 & 22,396 & 2,954 & 4,079 & 29,429   & evidence (23\%), no evidence (77\%) \\
\Xhline{2\arrayrulewidth}
\end{tabularx}
\caption{Splits, classes, and class distributions for all used datasets.}
\label{table_datasets_stats}
\end{table*}

\begin{table*}[!hbt]
\scriptsize
\centering 
\def\arraystretch{1.5}
\begin{tabularx}{\linewidth}{l L{1.8cm} L{3cm} L{6cm} l}
\Xhline{2\arrayrulewidth}
\textbf{Dataset} & \textbf{Domain}  & \textbf{Topic} & \textbf{Sentence} & \textbf{Class} \\\hline

\multirow[t]{ 2}{*}{\acs{fsc}} & \multirow[t]{ 2}{*}{Web Search} & electronic cigarettes & Currently, there is no scientific evidence confirming that electronic cigarettes help smokers quit smoking cigarettes. & contra \\
 &  & renewable energy & Installation is quick and homeowners can be enjoying solar energy in a matter of days. & pro \\
\hline
\multirow[t]{ 2}{*}{\acs{ukpc}} & \multirow[t]{ 2}{*}{Web Search} & nuclear energy & It is pretty expensive to mine, refine and transport uranium. & contra \\
 &  & gun control & Gun control laws would reduce the societal costs associated with gun violence. & pro \\
\hline
\multirow[t]{ 2}{*}{\acs{iam}} & \multirow[t]{ 2}{*}{Encyclopedia} & Should you restrict reality TV & They involve extreme competition which drains children; it takes away their innocence. & contest \\
 &  & Should boxing be banned & With a careful and thoughtful approach, boxing quite can be beneficial to health. & support \\
\hline
\acs{ibm} & Encyclopedia &  We should ban organic food & Like local food systems, organic food systems have been criticized for being elitist and inaccessible. & argument \\
\hline
\Xhline{2\arrayrulewidth}
\end{tabularx}   
\caption{All datasets used in this work with the general domain the data origins from and data samples with topic, sentence, and annotated labels (class).
}
\label{table_dataset_examples}
\end{table*}

\section{Data}
For our dataset composition benchmark, we first need two datasets that only differ in the aforementioned dimensions of \textit{number of topics} and \textit{number of samples per topic} but are otherwise similarly composed. We use one existing dataset (see Section \ref{sec:data_ukp}) and base a new dataset (see Section \ref{sec:data_fsc}) on it with a composition better fit for diversity sampling and few-shot learning. We evaluate our hypotheses on dataset composition for \ac{am} on two more \ac{am} datasets (see Sections \ref{sec:data_iam} and \ref{sec:data_ibm}) with slightly different learning tasks. Statistical information about all datasets as well as examples can be found in Tables \ref{table_datasets_stats} and \ref{table_dataset_examples}. Information about dataset licenses are listed in Appendix \ref{app:licenses}.

\subsection{\ac{ukpc}}\label{sec:data_ukp} As opposed to other \ac{am} datasets \cite{ibm-corpuswide,shnarch-etal-2018-will,rinott-etal-2015-show,aharoni-etal-2014-benchmark}, the \ac{ukpc} has two main advantages: First, it includes stance labels, which are an important additional information to categorize mined arguments and can be further processed for tasks like fake news detection \cite{hanselowski-etal-2018-retrospective}. Second, the dataset is from heterogeneous data sources and models real-world scenarios better than taking only samples from a single source. It consists of eight topics with a total of 25,492 samples, which are pairs of a short topic and a single sentence, labeled with \textit{argument for} (pro), \textit{argument against} (con), or \textit{no argument} (none). As described by \citet{Stab2018b}, a sentence is only labeled as pro or con argument, if it holds evidence for why the sentence supports or opposes the topic. If the sentence holds no such evidence or is unrelated to the topic, it is labeled as \textit{no argument}. We split the dataset by taking all samples of five topics for training, of one topic for development, and of two topics for testing. To allow for a fair comparison, we downsample the number of samples in the training set (equally for each topic) to fit the total number of training samples generated for our newly created \ac{fsc}. 

\subsection{\ac{fsc}}\label{sec:data_fsc} Due to its aforementioned advantages, we decide to base our new dataset on the \ac{ukpc}. We follow the exact guidelines and data crawling strategy used for the \ac{ukpc} and crowdsource 21,600 samples over 150 controversial topics with exactly 144 samples for each topic (see Appendix \ref{app:dataset_criteria}). The composition of our dataset is therefore ideal for few-shot learning, as we have the same number of samples for each topic to easily scale up and down from 0 to 144. Moreover, we have a large amount of topics to scale diversity up and down. The topics are a collection of controversial subjects from multiple domains like politics, technology, economy, and do not intersect with topics from the \ac{ukpc} (see Appendix Table \ref{tbl:dataset_topics_ukp_fs150t}). We randomly pick 10 of the topics for our development set (1,440 samples) and 20 topics for our test set (2,880) samples, leaving 120 topics for the training set (17,280 samples). 

\subsection{\ac{iam}}\label{sec:data_iam} The \ac{iam} is built upon the data from ``Task 1: Claim Extraction'' by \citet{cheng-etal-2022-iam}. The original data is based on 123 debating topics and 1,010 related articles from English Wikipedia. One sample consists of a topic and a sentence from an article. Each pair has one of three possible labels attached: \textit{support}, \textit{contest}, or no \textit{relation}. Due to the massive imbalance of the none-arguments in the original training split (93\%), we have to downsample them in a way that the model is able to pick up the other two classes. We randomly pick samples until we reach a class distribution of 22\%/18\%/60\% (support/contest/no relation) in the training set. We leave the dev and test sets untouched from the original, which also makes this dataset the only in-topic dataset (as opposed to cross-topic datasets which have no overlapping topics between the dataset splits). The modified training data set contains all 100 topics, the original dev and test sets contain 62 topics and 63 topics.

\subsection{\ac{ibm}}\label{sec:data_ibm} The \ac{ibm} is based on the publicly available dataset constructed by the authors in \citet{ibm-corpuswide}. The dataset consists of almost 30,000 \textit{motion}-sentence samples and each sample has a score between 0 and 1 that either denotes a sentence to be rather an evidence for the related motion or not. Motions are described as high-level claims, e.g. ``Capitalism brings more harm than good''. The sentences are extracted from English Wikipedia. Following the authors' experimental setup, we set a threshold at 0.6 for the score to define two class labels \textit{evidence} and \textit{no evidence}. We take 35 random topics to form the test set, 20 random topics to form the dev set, and 166 topics to form the training set. In contrast to all other datasets, this one has only two class labels and the largest number of topics (here: motions) and samples.

\section{Method}\label{sec:method}
To investigate the optimal composition of \ac{am} datasets, we conduct sample, topic, and dataset experiments, which we elaborate in the following.

\subsection{Sample experiments} We investigate on how many training samples per topic are necessary to reach \textit{acceptable} and maximum performance. We start our experiments with 0 training samples per topic, i.e. untrained model performance (zero-shot) and increase the number of samples in small steps, ending with all samples available for each topic. We define acceptable performance by reaching at least 95\% of the highest performance on a test set, measured over all sample experiments for a given model. We define maximum performance for a model by the highest value for the given metric on a test set, regardless of the number of training samples used to reach it.

\subsection{Topic experiments} We analyze how many topics are needed to generalize well in cross-topic experiments by choosing a set of training sample sizes (960; 1,440; 2,880; 5,760) and fixing them while increasing the topics. The topics are increased in steps of 5 and end with the maximum number of topics available for a training set. Since fixing the number of topics and sample sizes requires a certain amount of samples available for each topic in the training set, experiments with larger sample sizes may start at larger topic sizes. For instance, fixing 10 topics and 960 samples only requires 96 samples per topic in the training set, while fixing 10 topics and 5,760 samples already requires 576 samples per topic. The more topics we include for a fixed sample size, the less samples per topic are available. We aim to find out whether or not using many topics (diversity sampling) is beneficial for cross-topic performance.

\subsection{Dataset experiments} We investigate if we can reach higher performance by training on a dataset with few topics but many samples per topic (\ac{ukpc}) or on a dataset with many topics but few samples per topic (\ac{fsc}). 
For the benchmark experiments, we leverage both supervised models and re-train them in the following to setups:
\begin{itemize}[noitemsep,topsep=0pt]
    \item Training and tuning on the \ac{ukpc} and show results on the test sets of both corpora.
    \item Training and tuning on the \ac{fsc} and show results on the test sets of both corpora.
\end{itemize}
If either of the two variants performs better in both experiments, we know the dataset composition that should be preferred for the task of \ac{am}. In any other case, our assumption that few samples combined with many topics is the superior dataset composition (i.e. better cross-topic performance) on \ac{am} datasets is refuted.

\section{Models}
We use four models: ERNIE 2.0 as a strong and fast to train language model, FLAN-T5 XL as an LLM option that was trained on massive amounts of data, and two state-of-the-art chat models for zero-shot in-context learning experiments. Details on fine-tuning parameters are described in Appendix \ref{app:hyperparam}. 

\subsection{ERNIE 2.0} As medium-sized model (110M parameters), we use ERNIE 2.0 \cite{ernie_aaai} which was pre-trained in a continual multi-task learning fashion on several word-, structure-, and semantic-aware tasks (but not on \ac{am} tasks) and showed state-of-the-art performance when fine-tuned on tasks of the GLUE Benchmark \cite{wang2019glue}. The data for the pre-training tasks was automatically generated with text extracted from encyclopedias, books, dialog, and discourse relation datasets. As these tasks have similar properties to \ac{am}, we expect to benefit from the pre-training through a higher maximum performance. Moreover, we anticipate that the specific pre-training enables the model to bootstrap performance on few-shot learning. 

\subsection{FLAN-T5 XL} We use FLAN-T5 XL \cite{https://doi.org/10.48550/arxiv.2210.11416} as a \textit{large} language model for our experiments. It is a variant of the T5 model \cite{10.5555/3455716.3455856} which was fine-tuned on 1.8K instruction tasks. As this model has an encoder-decoder structure, we remove the decoder and use a classification head for our tasks, which leaves around 1.3B parameters. Fine-tuning is done with LoRa \cite{hu2022lora} to reduce training time.

\subsection{LLama2-70B, ChatGPT} As strong, zero-shot baselines, we use Llama2-70B \cite{touvron2023llama} and ChatGPT \cite{chatgpt35} in our experiments. The prompts used for each dataset and the specific model versions can be found in Appendix \ref{app:hyperparam_llm}.

\section{Topic \& Sample Experiments and Evaluation}
We run all experiments over six seeds and report the average F$_1$ macro on the test set for the three seeds with the highest F$_1$ macro measured on the development set (see Appendix \ref{app:hyperparam}).\footnote{With low sample sizes, the models sometimes fail on some of the seeds and distort the results.}

\subsection{Sample Experiments}\label{sec:sample_exp}
\begin{figure}
  \centering
  \includegraphics[width=1.0\linewidth]{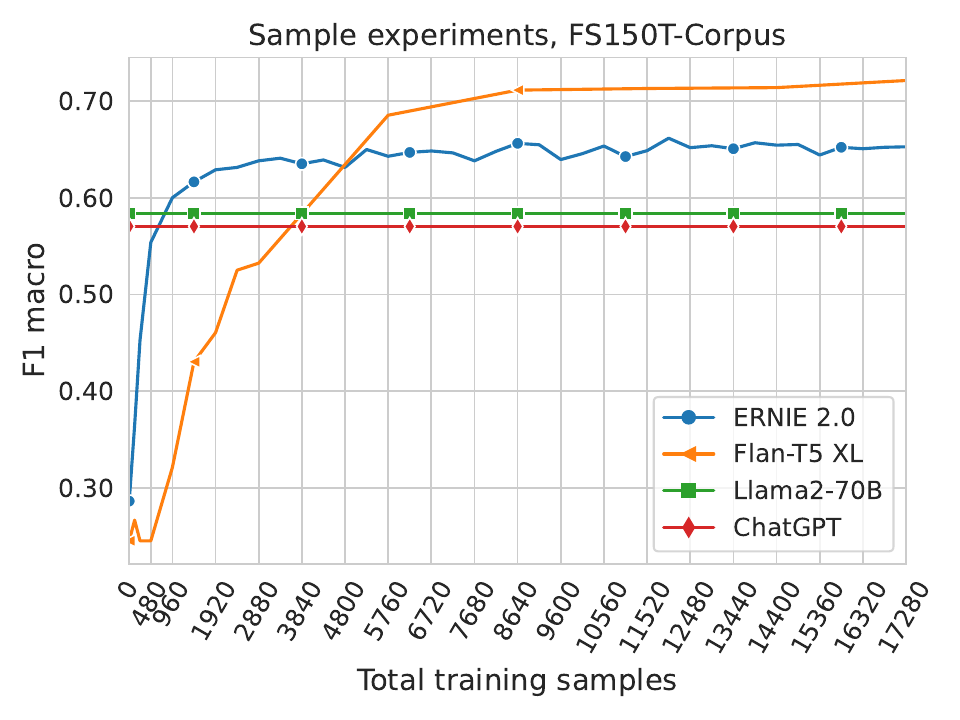}
  \caption{Sample experiments on the \ac{fsc}}
  \label{fig:sample_experiment_plot_fsc}
\end{figure}

\begin{figure}
  \centering
  \includegraphics[width=1.0\linewidth]{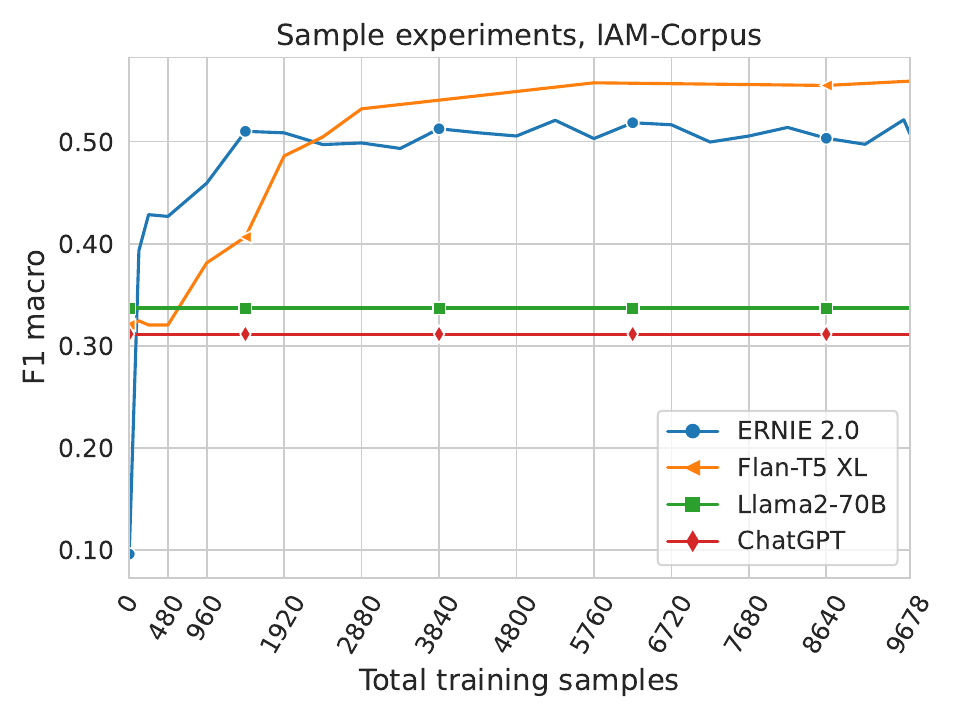}
  \caption{Sample experiments on the \ac{iam}}
  \label{fig:sample_experiment_plot_iam}
\end{figure}

\begin{figure}
  \centering
  \includegraphics[width=1.0\linewidth]{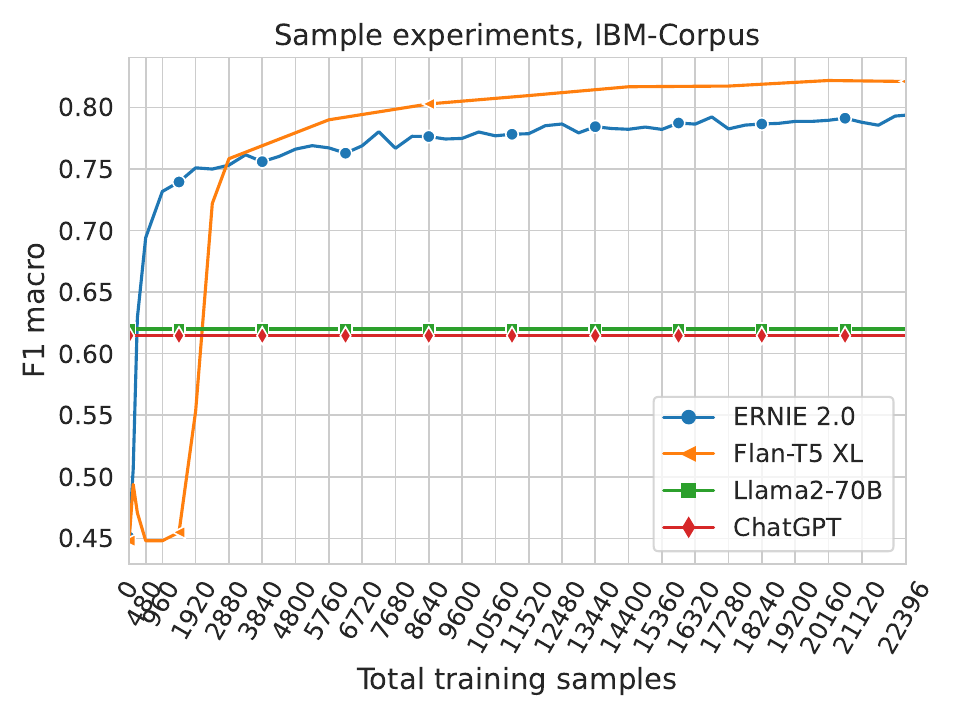}
  \caption{Sample experiments on the \ac{ibm}}
  \label{fig:sample_experiment_plot_ibm}
\end{figure}

Figures \ref{fig:sample_experiment_plot_fsc}-\ref{fig:sample_experiment_plot_ibm} (see also Appendix \ref{app:sample_topic_figures_err}) show the performance gains of all tested models with increasing sample sizes per topic (samples uniformly distributed over all training topics)\footnote{Increased step size of 2,880 for FLAN-T5 XL from 2,880 samples onwards.}. As two strong zero-shot baselines, we show results with Llama2-70B and ChatGPT.

For all datasets, we observe that FLAN-T5 XL struggles with small sample sizes, whereas ERNIE 2.0 shows good performance early on. ERNIE 2.0 reaches $>$0.60 F$_1$ macro on the \ac{fsc} at 960 samples (FLAN-T5 XL: $>$2,800 samples), $>$0.70 F$_1$ macro on the \ac{ibm} at 960 samples (FLAN-T5 XL: 2,400 samples), and $>$0.50 F$_1$ macro on the \ac{iam} at 1,440 samples (FLAN-T5 XL: 2,400 samples). However, when looking at maximum performance, FLAN-T5 XL eventually outperforms ERNIE 2.0 on all datasets on up to 6pp (percentage points) in F$_1$ macro. Both models require all training data to reach their maximum performance, except ERNIE 2.0 on the \ac{fsc} where it peaks at 69\% of the training data and FLAN-T5 XL on the \ac{ibm} where it still requires 90\% of the training data. Llama2-70B and ChatGPT outperform the other models on all datasets on zero-shot experiments. However, they both loose their advantage after 500-2,000 training samples eventually, depending on the dataset. Interestingly, on our \ac{fsc} that is especially designed for few-shot learning, both zero-shot baselines are the most competitive to our supervised learning models.


\begin{table}\small
\centering 
\resizebox{0.5\textwidth}{!}{
\def\arraystretch{1.3}
\begin{tabular}{L{1.0cm} L{1.8cm} C{1.5cm} C{2.0cm} c}
\Xhline{2\arrayrulewidth}
Dataset & Model  & \makecell{\# Samples:\\ max performance} & F$_{1}$ macro & \makecell{\# Samples: \\95\% of max \\ performance} \\\hline
\multirow[c]{5}{*}{\makecell{FS150T-\\Corpus}} 

      & FLAN-T5 XL  & 17,280 (100\%) & .7214 $\pm$ .0064 & 5,760 (33\%)  \\
       
      & ERNIE 2.0 & \textbf{12,000} (69\%) & .6617 $\pm$ .0048 & \textbf{1,920} (11\%) \\

      & ChatGPT & 0 & .5700 & - \\

      & Llama2-70B & 0 & .5608 & - \\

      & Majority & 0 & .2451 & - \\

      \hline

\multirow[c]{5}{*}{\makecell{IAM-\\Corpus}} 
      & FLAN-T5 XL  & 9,678 (100\%) & \textbf{.5591} $\pm$ .0234 & 2,880 (30\%)  \\
      
      & ERNIE 2.0 & \textbf{6,720} (99\%) & .5213 $\pm$ .0070 & \textbf{1,440} (15\%)\\
       
      & ChatGPT & 0 & .2890 & - \\

      & Llama2-70B & 0 & .2920 & - \\
       
      & Majority & 0 & .3204 & - \\

      \hline

\multirow[c]{5}{*}{\makecell{IBM-\\Corpus}} 

      & FLAN-T5 XL  & \textbf{20160} (90\%) & \textbf{.8210} $\pm$ .0043 & 5760 (26\%)  \\
       
      & ERNIE 2.0 & 22,396 (100\%)& .7937 $\pm$ .0015 & \textbf{3,360} (15\%) \\

      & ChatGPT & 0 & .6150 & - \\

      & Llama2-70B & 0 & .6210 & - \\
       
      & Majority & 0 &  .4481 & - \\

       \Xhline{2\arrayrulewidth}
\end{tabular}
}
\caption{Sample experiment results with training samples required for highest performance, highest performance in F$_{1}$ macro, and number of training samples required to reach 95\% of the highest performance.
}\label{tbl:model_smp_performance}
\end{table}

\begin{figure*}
\centering
\begin{subfigure}{0.33\textwidth}
  \centering
  \includegraphics[width=1.0\linewidth]{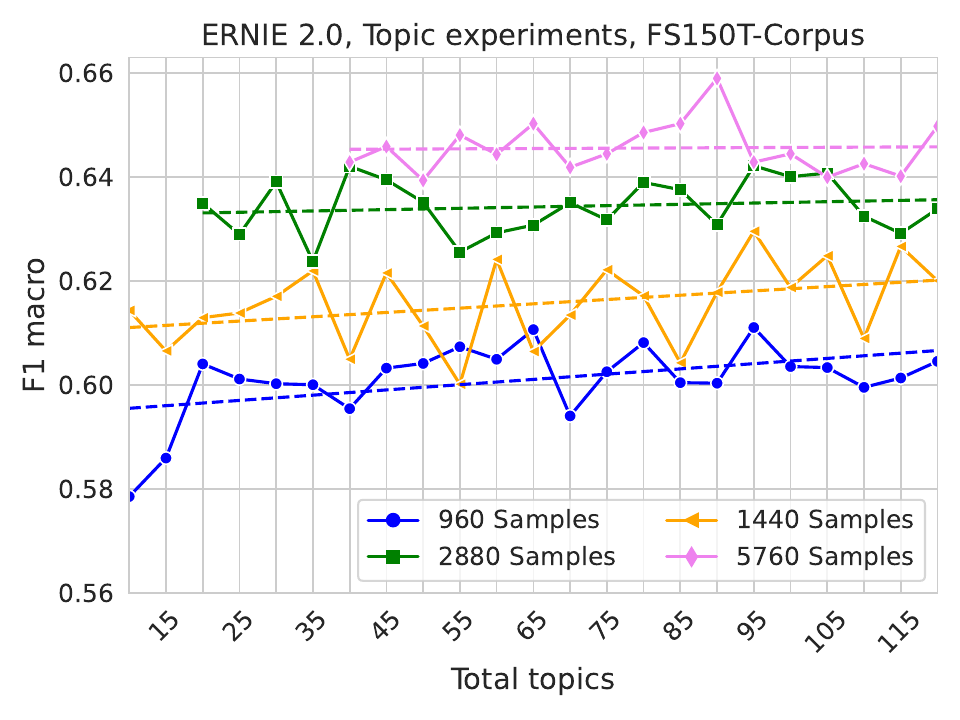}

  \label{fig:topic_experiment_plot_fsc_ernie}
\end{subfigure}%
\begin{subfigure}{0.33\textwidth}
  \centering
  \includegraphics[width=1.0\linewidth]{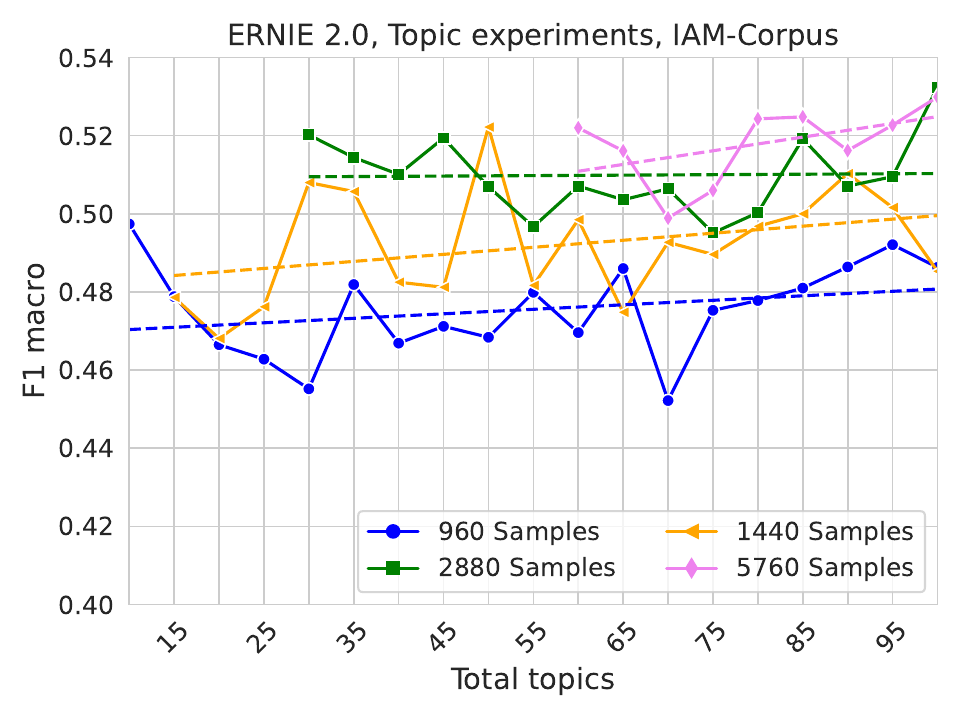}

  \label{fig:topic_experiment_plot_iam_ernie}
\end{subfigure}
\begin{subfigure}{0.33\textwidth}
  \centering
  \includegraphics[width=1.0\linewidth]{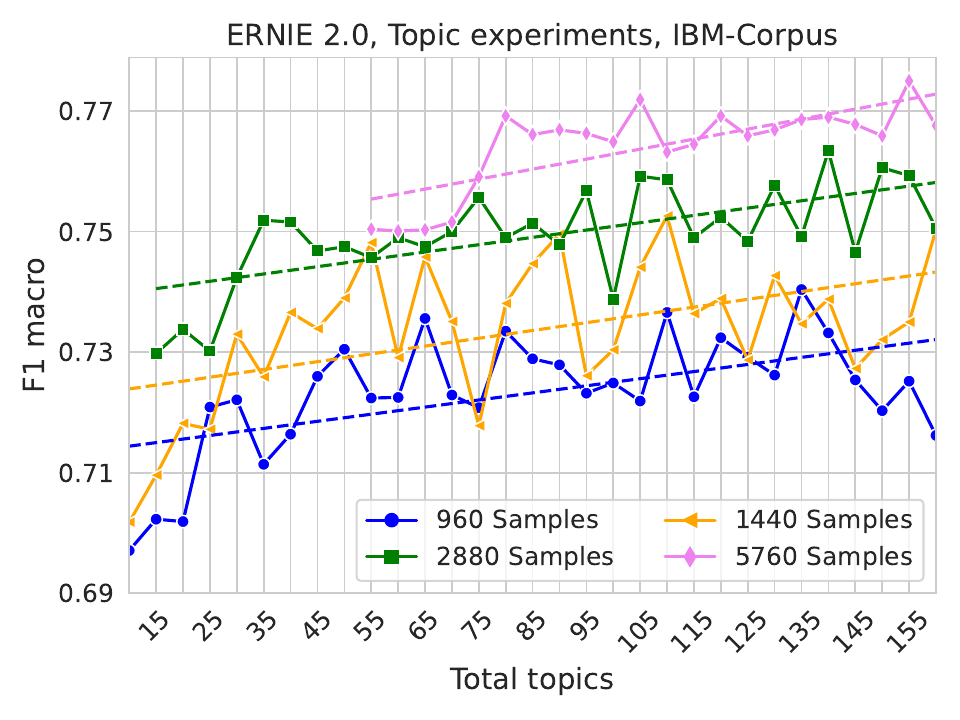}

  \label{fig:topic_experiment_plot_ibm_ernie}
\end{subfigure}

\begin{subfigure}{0.33\textwidth}
  \centering
  \includegraphics[width=1.0\linewidth]{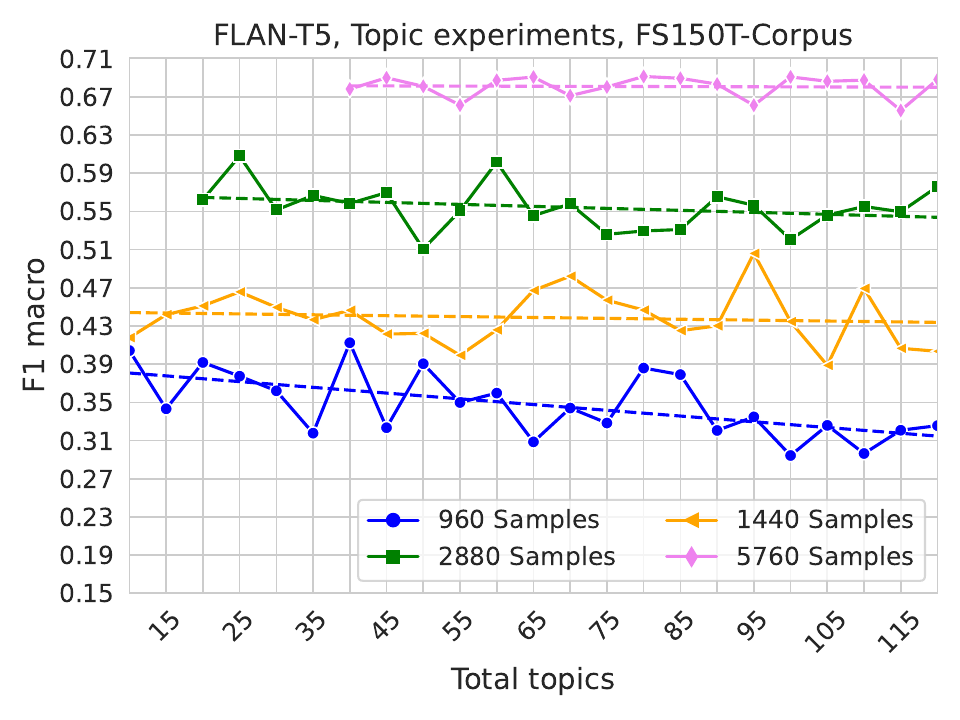}
  \label{fig:topic_experiment_plot_fsc_flan}
\end{subfigure}%
\begin{subfigure}{0.33\textwidth}
  \centering
  \includegraphics[width=1.0\linewidth]{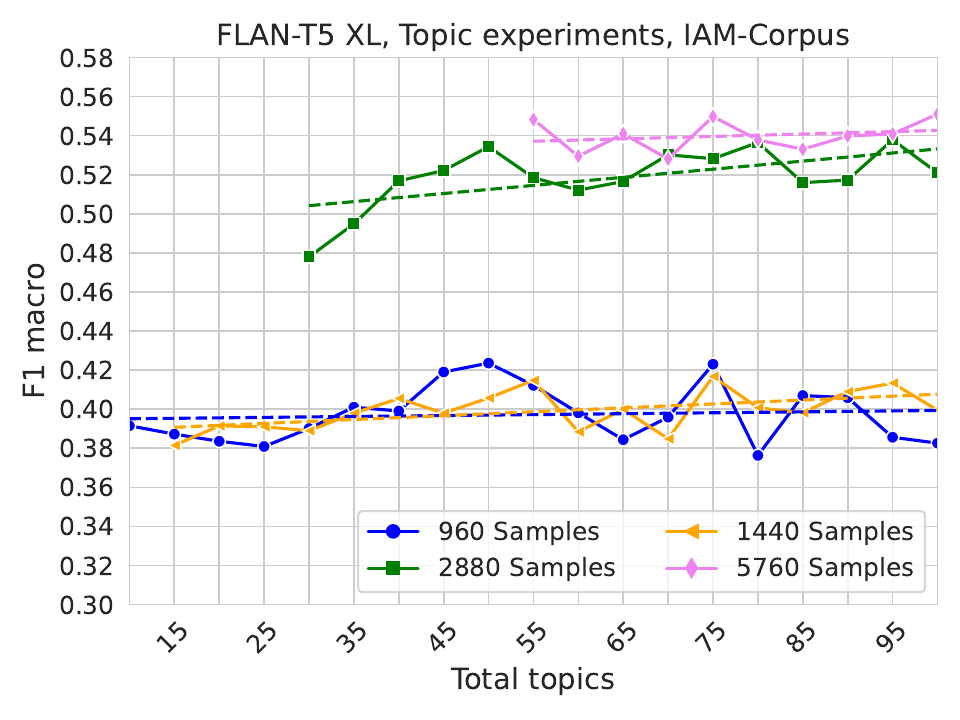}
  \label{fig:topic_experiment_plot_iam_flan}
\end{subfigure}
\begin{subfigure}{0.33\textwidth}
  \centering
  \includegraphics[width=1.0\linewidth]{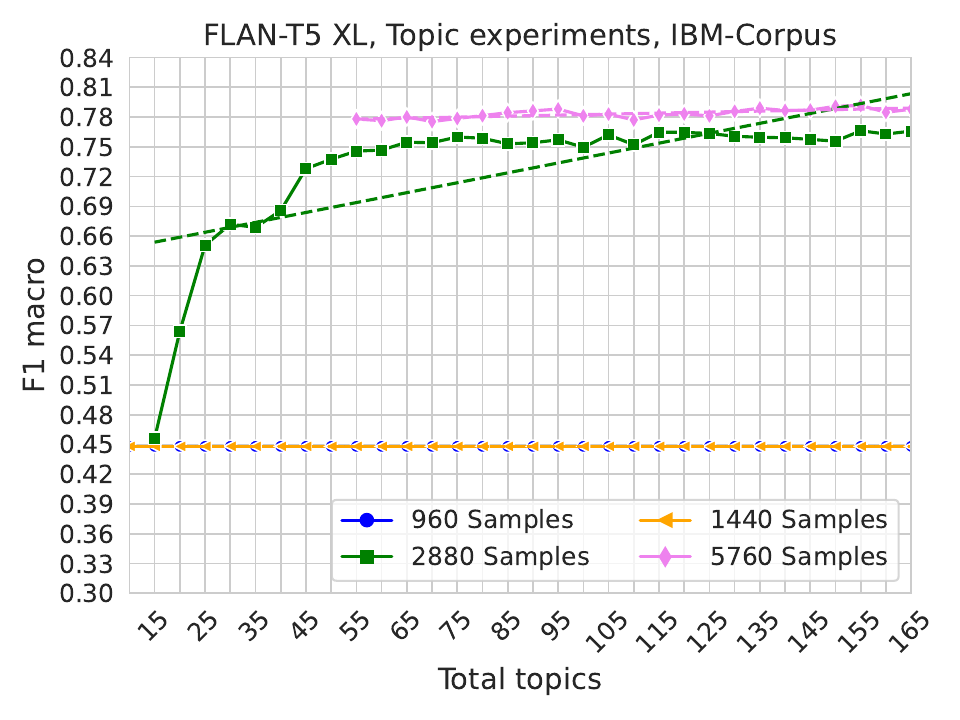}
  \label{fig:topic_experiment_plot_ibm_flan}
\end{subfigure}

\caption{Topic experiments for FS150T-/IAM- and IBM-Corpus on ERNIE 2.0 and FLAN-T5 XL and in F${_1}$ macro.}\label{fig:topic_experiment_plot}
\end{figure*}

We also investigate how much data we really need to reach \textit{acceptable} performance, i.e., 95\% of the maximum performance of a model (see Table \ref{tbl:model_smp_performance}). For all three datasets, ERNIE 2.0 only needs a maximum of 15\% of the data to reach this performance. For the \ac{fsc}, it only needs 11\% (with as few as 16 samples per topic). Hence, when using ERNIE 2.0, we can almost drop 90\% of the data (if the dataset has a composition as proposed in this work). For FLAN-T5 XL, which has difficulties with smaller sample sizes on the datasets, it still only needs 26-33\% to reach 95\% of its maximum performance on the datasets, which could also reduce the necessary data size by almost 70\%.

\subsection{Topic Experiments}\label{sec:topic_exp}
For the experiments, we fine-tune ERNIE 2.0 and FLAN-T5 XL on four different training sample sizes from 960 to 5,760. Topics and samples for each run are picked randomly.

We have a deeper look into how model performance changes for all datasets if the number of topics is increased while the training sample size is fixed (see Figure \ref{fig:topic_experiment_plot} and Appendix \ref{app:sample_topic_figures_err}). By increasing the number of topics, we also increase the diversity of the training set through adding outliers, instead of just picking more samples with similar content from the existing topics.

Similar to the increased robustness observed by \citet{larson-etal-2019-outlier}, ERNIE 2.0 shows an upward trend (dashed lines) for all datasets and sample sizes. For the \ac{fsc}, we observe an upward trend with up to to 1.1pp on 960 samples, decreasing when more training samples are used. The largest upward trend for \ac{iam} is reached on 5.760 samples (2.2pp) and for the \ac{ibm}, the largest upward trend is reached on 1,440 samples (1.9pp). Hence, ERNIE 2.0 is able to leverage diverse topic distribution in all tested scenarios.
For the much larger FLAN-T5 XL, the impact of topic diversity is mixed. For experiments with 1,440 and less samples, we observe either no clear trend or even a negative trend (also slightly for 2,880 samples on the \ac{fsc}). Using more samples, we observe a more positive trend. We assume these mixed results are due to two reasons: First, FLAN-T5 XL has seen much larger quantities of data initially in its pre-training as compared to ERNIE 2.0, which makes it harder to add even more diverse training data. Second, the large model size leads to unstable results with low sample sizes---we observe this for the \ac{ibm} in the topic experiments and, generally, in the sample experiments (see Figures \ref{fig:sample_experiment_plot_fsc}-\ref{fig:sample_experiment_plot_ibm}). Hence, we conclude that the significance of the experiments with FLAN-T5 XL on smaller sample sizes is low. 

\section{Cross-Dataset Experiments and Evaluation on Benchmark Dataset}\label{sec:datasetExp}

We tune two models based on ERNIE 2.0 and FLAN-T5 XL: one model trained and tuned on the \ac{fsc} and one model trained and tuned on the \ac{ukpc}. We show evaluation results for both models on both corpora in Table \ref{tbl:model_dataset_performance}.

Our baseline setting is when training, tuning, and testing happens on the same dataset. 
For that setup, FLAN-T5 XL shows .7532 and .7343 F${_1}$ macro for the \ac{ukpc} and the \ac{fsc}, respectively. Training and tuning FLAN-T5 XL on the \ac{ukpc} and then evaluating it on the \ac{fsc} shows worse performance (.6917 F${_1}$ macro) than training and tuning a model on the actual \ac{fsc}, which is the expected outcome. However, training and tuning FLAN-T5 XL on the \ac{fsc} shows the best results on the \ac{ukpc} test set (.7944 F${_1}$ macro). The same observation is made with ERNIE 2.0, just with lower overall F${_1}$ macro scores.  Hence, using the \ac{fsc} for training performs best on the test sets of both corpora. We assume that training on just a few topics and fitting the model with a large number of training samples to those topics will not prepare it enough to generalize well --- not even for a massively pre-trained model like FLAN-T5 XL. Training on many diverse topics, however, will add generalizability to the model and reduce the risk to over-fit to a small range of specific topics, that is, it will also learn that topics can come from a  much larger variety within the embedding space.

We also compare the models' performances to the current state-of-the-art (for which topic-wise results are available) on the \ac{ukpc}. TACAM-BERT Base \cite{tacam2019}, with a number of parameters comparable to ERNIE 2.0, performs 21.6pp and 5.1pp lower in F$_1$ macro for the test topics \textit{minimum wage} and \textit{school uniforms}. The much larger TACAM-BERT Large (three times the number of parameters) still underperforms ERNIE 2.0 by 1.6pp and 5.1pp.

\begin{table}
\centering 
\resizebox{0.48\textwidth}{!}{
\def\arraystretch{1.3}
\begin{tabular}{cl|ccc|c}
\Xhline{2\arrayrulewidth}
 & & \multicolumn{3}{c}{\makecell{Test on \\ \ac{ukpc}}} &  \makecell{Test on \\ \ac{fsc}} \\
 
 & \makecell[r]{Test topics:} & MW & SU & all & all \\
 
 \hline
 
\makecell[c]{TACAM-\\BERT Base\textsuperscript{*}} & \makecell[l]{Train \& tune on \\ \ac{ukpc}} & .4900 & .6900 & - & - \\
\makecell[c]{TACAM-\\BERT Large\textsuperscript{*}} & \makecell[l]{Train \& tune on \\ \ac{ukpc}} & .6900 & .6900 & - & - \\

\hline

\multirow{3}{*}{\makecell[c]{ERNIE\\2.0}} & \makecell[l]{Train \& tune on \\ \ac{ukpc}} & \makecell{.6777 \\$\pm$ .0118} & \makecell{.7149 \\$\pm$ .0093} & \makecell{.6980 \\$\pm$ .0085} & \makecell{.6292 \\$\pm$ .0072} \\

 & \makecell[l]{Train \& tune on \\ \ac{fsc}} & \makecell{\textbf{.7058 } \\$\pm$ .0046}& \makecell{\textbf{.7406} \\$\pm$ .0029} & \makecell{\textbf{.7243} \\$\pm$ .0048} & \makecell{\textbf{.6585} \\$\pm$ .0036}  \\

\hline

\multirow{3}{*}{\makecell[c]{FLAN-T5\\XL}} & \makecell[l]{Train \& tune on \\ \ac{ukpc}} & \makecell{.7333 \\$\pm$ .0095} & \makecell{.7881 \\$\pm$ .0114} & \makecell{.7532 \\$\pm$ .0132} & \makecell{.6917 \\$\pm$ .0132} \\

 & \makecell[l]{Train \& tune on \\ \ac{fsc}} & \makecell{\textbf{.7574 } \\$\pm$ .0055} & \makecell{\textbf{.8270} \\$\pm$ .0023} & \makecell{\textbf{.7944} \\$\pm$ .0021} & \makecell{\textbf{.7343} \\$\pm$ .0037} \\

\Xhline{2\arrayrulewidth}
\end{tabular}
}
\caption{Dataset experiment results with ERNIE 2.0 and FLAN-T5 XL, comparing results on the \ac{fsc} and \ac{ukpc}. As a baseline for the \ac{ukpc}, we use TACAM-BERT (\textsuperscript{*}work by \citet{tacam2019}). MW=Minimum Wage, SU=School uniforms.}\label{tbl:model_dataset_performance}
\end{table}

\section{Conclusion}\label{sec:conclusion}
We create a new dataset that enables to benchmark the composition of \ac{am} datasets. Experiments show that having many topics in combination with few samples per topic can improve model performance by 4.1pp in cross-dataset experiments and also reaches a new state-of-the-art on the \ac{ukpc} (see Section \ref{sec:datasetExp}).

\textbf{Recommendations for dataset composition:} Overall, we observe a positive trend in performance when the number of training topics is increased and a medium-sized LM is used (ERNIE 2.0), but mixed results with an LLM (FLAN-T5 XL), which we attribute to the extensive and diverse pre-training and generally more unstable results on smaller sample sizes (see Section \ref{sec:topic_exp}). While the topic experiments do not show a drastic increase of accuracy, it can be an easy way to improve the performance and usually comes without additional costs. Hence, if there is a sample limit for a planned dataset, we can increase a medium-sized model's performance by composing the dataset with more topics. As we tested up to 160 topics on all datasets, we assume this to be a good choice for training data sizes ranging from \textasciitilde1,000-6,000 samples but can become less relevant if a large model like FLAN-T5 XL is used. However, in many scenarios where inference speed and operating costs are decisive, a smaller model like ERNIE 2.0 with carefully sampled training data might be the preferred choice.\footnote{In comparison on an NVIDIA A10, FLAN-T5 XL can predict 34 samples per second with a batch size of 1 and takes 5,638MB of GPU memory, whereas ERNIE 2.0 can predict 111 samples per second and only takes 860MB.} 

When choosing our proposed dataset composition (\ac{fsc}) in combination with ERNIE 2.0 (pre-trained on several word-, structure-, and semantic-aware tasks), we can reduce the training sample size by almost 90\% (to 1,920 samples), still reach 95\% of the maximum performance and, in turn, decrease the annotation costs of the train set by \$2,323 to only \$290 (see Appendix \ref{app:dataset_criteria}) for a dataset created in the composition proposed in this work (see Section \ref{sec:sample_exp}). 
Although showing difficulties on small samples sizes, we can still reduce the sample size by 67\% with FLAN-T5 XL. We observe the same trend on the other two datasets: on the \ac{ibm}, 15\% of the training data with ERNIE 2.0 and 26\% with FLAN-T5 XL are sufficient to reach 95\% of the maximum performance; on the \ac{iam}, 15\% with ERNIE 2.0 and 30\% with FLAN-T5 XL are sufficient. This clearly challenges the trend to develop larger datasets for \ac{am}. Following our proposed dataset composition makes low-budget productions of high-quality \ac{am} datasets possible, contributing to a more diverse landscape of those datasets. 

\textbf{Generalization to other tasks:} While we test our approach on \ac{am} only, work on related tasks also shows performance improvements when focusing on dataset diversity. For instance, \citet{arakelyan-etal-2023-topic} show that higher performance can be reached if subsets of stance detection datasets are sampled for diversity with an unsupervised topic model and used for training. \citet{sultan-etal-2020-importance} use a transformer-based question generator and conclude that more diverse questions lead to a higher performance on downstream Question-Answering tasks. \citet{yadav-etal-2024-explicit} also show improvements for Question-Answering by generating question-answer pairs with a focus on diversity conditions like spacial aspects, question types and entities. 
Similar to our findings, they also show that performance gains are highest in low-resource scenarios. 
Hence, we speculate that a well-architected dataset composition, based on diversity, can also lead to a high performance on Question-Answering and pure Stance Detection datasets.

We publish our newly created dataset\footnotemark[1] and code\footnotemark[2], allowing for further benchmarking experiments to develop the design of future \ac{am} datasets.

\section*{Limitations}

Our experiments focus on datasets for \ac{am} only. While we would expect other tasks with datasets of similar composition (for instance, Stance Detection or Question-Answering, as discussed in Section \ref{sec:conclusion}) to also profit from our findings, we have not tested this and can only make claims based on our experiments for \ac{am}. 

Our sample experiments only cover sample sizes from 0 to 22,396 training samples and a step size of 480-2,880 samples, depending on the model. Hence, we can not rule out the possibility of higher F$_1$ macro or other derivations from the observable trend with more than 22,396 training samples, nor can we rule out the possibility that we have missed a certain dip or peak due to our chosen step size. 
Similarly for our topic experiments---while there is a trend to higher performance with more topics, it is unclear how this trend develops with more than 160 topics (for instance, if the model shows a saturation with regard to topics or if more topics would even have a negative impact). 

Lastly, when comparing the maximum performance of our medium sized model choice with our LLM choice, there is an obvious gap in performance to observe. There is, however, a clear downside of using LLMs when it comes to operation cost and speed (see Section \ref{sec:conclusion}), which can be a crucial factor in many scenarios.

\section*{Acknowledgements}
This work has been funded by the LOEWE Distinguished Chair “Ubiquitous Knowledge Processing”, LOEWE initiative, Hesse, Germany (Grant Number: LOEWE/4a//519/05/00.002(0002)/81) and by the German Federal Ministry of Education and Research (BMBF) under the promotional reference 01UP2229B (KoPoCoV).

\bibliography{custom_rebiber}

\appendix
\section{Reproducibility Criteria}

\subsection{Dataset}\label{app:dataset_criteria} The new dataset \ac{fsc} consists of 21,600 samples over 150 controversial topics with 144 samples each. We index the CommonCrawl\footnote{\url{https://commoncrawl.org}} dump \textit{CC-MAIN-2016-07} via ElasticSearch\footnote{\url{https://www.elastic.co}} and use all 150 controversial topics to search and extract texts for the crowdsourcing process. To split the texts into sentences, we use NLTK 3.7 \cite{bird2009natural}. The topics for the dataset are a collection of controversially discussed subjects from the domains of, amongst others, politics, technology, and economy. They were gathered manually from twitter and reddit trends, as well as various discussion forums. See Table \ref{tbl:dataset_topics} for a list of all topics in alphabetical order, including the semantically closest topic for each given in cosine similarity. We compute the similarity of two topics by averaging the pairwise cosine similarity of all sentences for two topics. The embeddings were generated by using the sentence-transformer library\footnote{\url{https://github.com/UKPLab/sentence-transformers}} (version 2.2.2) \cite{reimers-gurevych-2019-sentence}. As the highest cosine similarity between two topics is only 0.25, this indicates low overlapping in general.

We split the dataset into a \textit{train}, \textit{development}, and \textit{test} set. There is no overlap between topics of the sets or with topics of the \ac{ukpc} (see Table \ref{tbl:dataset_topics_ukp_fs150t}). The dataset language is English and the annotation guidelines for the crowdsourcing process are taken from \citet{Stab2018b}. See Tables \ref{table_datasets_stats} and \ref{table_dataset_examples} for more statistics and examples about the dataset. 

The crowdsourcing costs on Amazon Mechanical Turk\footnote{\url{https://www.mturk.com/}} amount to a total of \$3,266. The study was open to all people located in the US and we paid well above the US federal minimum wage of \$7.25/hour. Each sample was annotated by seven independent, anonymous annotators. We asked the annotators to label each sample they were presented with (consisting of the guidelines, a topic, and a respective sentence) into categories \textit{pro}, \textit{contra}, or \textit{none}. We design our guidelines based on \citet{Stab2018b}, i.e. a sentence is only to be labeled as \textit{pro} or \textit{contra} argument, if it holds evidence for why the sentence supports or opposes the topic. If the sentence holds no such evidence or is unrelated to the topic, it is labeled as no argument (\textit{none}). To generate gold labels, we apply the MACE denoising tool \cite{DBLP:conf/naacl/HovyBVH13} with a threshold of 0.9 as done in \cite{Stab2018b}. Finally, two experts were asked to annotate 100 randomly picked samples from the dataset. We create gold labels in the same way as for the crowdworker annotations. The Cohen's $\kappa$ between expert and crowdworkers is .52, which can be interpreted as ``moderate'' agreement \cite{10.2307/2529310} and reasonable for the complexity of the task and the large amount of different and difficult to understand topics.

\subsection{Models}\label{app:hyperparam} 

\subsubsection{ERNIE 2.0, FLAN-T5 XL} 
\begin{figure*}
\centering
\begin{subfigure}{0.49\textwidth}
  \centering
  \includegraphics[width=1.0\linewidth]{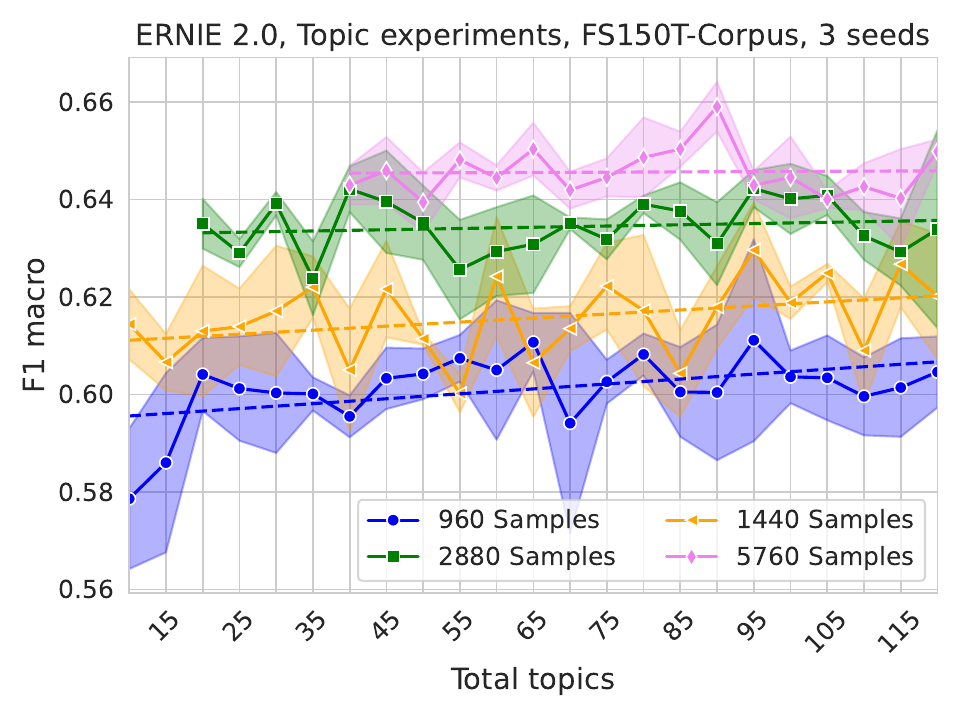}
  \label{fig:topic_exp_3v6_seeds_3_stdev}
\end{subfigure}%
\hfill
\begin{subfigure}{0.49\textwidth}
  \centering
  \includegraphics[width=1.0\linewidth]{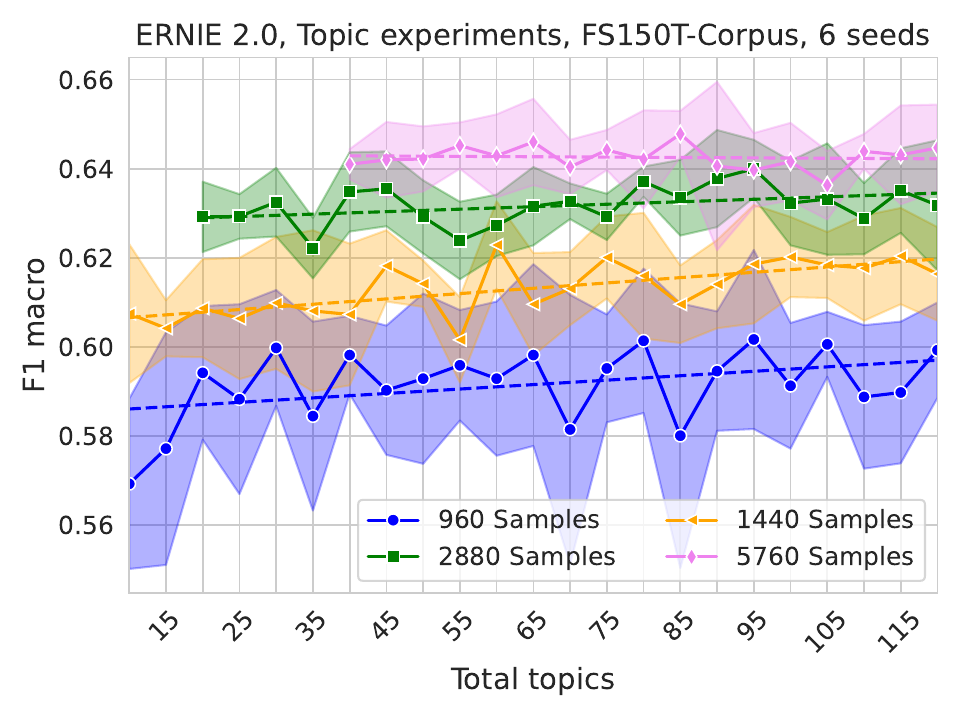}
  \label{fig:topic_exp_3v6_seeds_6_stdev}
\end{subfigure}

\caption{Same topic experiments with ERNIE 2.0 on \ac{fsc}, but taking only the best 3 seeds on the development set (left figure) or taking all 6 seeds (right side) into account. Using 6 seeds shows higher standard deviation and lower performance, especially for smaller sample sizes.}
\label{fig:topic_exp_3v6_seeds_stdev}
\end{figure*}

We tune both models on the full training sets with all combinations of four different learning rates ($1*10^{-4}$, $1*10^{-5}$, $3*10^{-5}$, $5*10^{-5}$) and three batch sizes (4, 8, 16). All models are trained over 5 epochs and we use the best model (always determined on the development set by highest F$_1$ macro) to fix the hyperparameters for the actual experiments. Due to unstable performance on low sample sizes, we decide to always train on 6 different seeds (for tuning and actual experiments), but only leverage the averaged results on the best 3 of them. 
To better understand the reasoning behind this approach, we show the difference of using the best 3 seeds or all 6 seeds on the topic experiments for the \ac{fsc} with ERNIE 2.0 (see Figure \ref{fig:topic_exp_3v6_seeds_stdev}). As can be seen, using 6 seeds shows an overall higher standard deviation and lower performance, especially for lower sample sizes.
The final hyperparameters found for each dataset and model are listed in Table \ref{tbl:models_hyperparameters}. Training ERNIE 2.0 (110M parameters) takes approx. 25 minutes on a single NVIDIA P-100 (one seed) with the full training set of the \ac{fsc} and approx. 45 minutes for our FLAN-T5 XL encoder (1.3B parameters) on a single NVIDIA A10 with LoRa rank 16. We use PyTorch 2.1.0 \cite{NEURIPS2019_9015} and transformers 4.37.1 \cite{wolf-etal-2020-transformers} to run the models and scikit-learn 0.23.2 \cite{scikit-learn} to compute the metrics.

\begin{table}\small
\centering 
\resizebox{0.48\textwidth}{!}{
\def\arraystretch{1.3}
\begin{tabular}{l|cc|cc}
\Xhline{2\arrayrulewidth}
 & \multicolumn{2}{c}{\makecell{ERNIE 2.0}} &  \multicolumn{2}{c}{\makecell{FLAN-T5 XL}} \\
 
 & Learning Rate & Batch Size & Learning Rate & Batch Size  \\
 
 \hline
 
\ac{fsc} & $1*10^{-5}$ & 4 & $1*10^{-4}$ & 4 \\
\ac{iam} & $1*10^{-5}$ & 16 & $1*10^{-4}$ & 4 \\
\ac{ibm} & $1*10^{-5}$ & 8 & $1*10^{-4}$ & 16 \\

\Xhline{2\arrayrulewidth}
\end{tabular}
}
\caption{Best hyperparameters for ERNIE 2.0 and FLAN-T5 XL on all datasets.}\label{tbl:models_hyperparameters}
\end{table}

\subsubsection{Llama2-70B-Chat, ChatGPT}\label{app:hyperparam_llm} We use Llama2-70B-Chat \cite{touvron2023llama} in a 4bit quantized version\footnote{\url{https://huggingface.co/TheBloke/Llama-2-70B-chat-AWQ/commit/ad4d622cb488138748dd28a0ca95c2b34dbe3964}} that we run on four NVIDIA A6000 with vLLM \cite{kwon2023efficient} and ChatGPT (gpt-3.5-turbo, September 25 Version) \cite{chatgpt35} via the OpenAI API\footnote{\url{https://platform.openai.com/docs/api-reference}}. We defined prompts that closely resemble the definitions for the respective dataset (for \ac{fsc}, the definition from \citet{Stab2018b} is used) and list all of them in Table \ref{table_dataset_llm_phrases}.

\begin{table*}[!hbt]
\scriptsize
\centering 
\def\arraystretch{1.5}
\begin{tabularx}{\linewidth}{L{1.6cm} C{4.3cm} C{4.3cm} C{4.3cm}}
\Xhline{2\arrayrulewidth}
\textbf{Dataset} & \textbf{Prompt 1}  & \textbf{Prompt 2} & \textbf{Prompt 3} \\\hline

\acs{fsc} & Decide if the below sentence is an argument with regard to the given topic. We define an argument as a span of text expressing evidence or reasoning that can be used to either support or oppose a given topic. An argument need not be ``direct'' or self-contained—it may pre-suppose some common or domain knowledge, or the application of commonsense reasoning—but it must be unambiguous in its orientation to the topic. If it is no argument, label it neutral. If it is an argument, decide whether it is in favor or against the topic and label it with favor or against. Only answer with neutral, favor, or against.
Sentence: [A sentence from the test set] 
Topic: [A respective topic from the test set] 
Label: 
&  
Decide if the below sentence is a pro argument (label it ``pro") or a contra argument (label it ``contra") regarding the given topic. If it is no argument regarding topic or no argument at all, label it "none":
Sentence: [A sentence from the test set] 
Topic: [A respective topic from the test set] 
Label:
& 
Label the sentence ``[A sentence from the test set]'' with ``pro'' if it is an argument regarding topic ``[A respective topic from the test set]'' or ``contra'' if it is an argument against the topic. Label it ``none'' if the sentence is no argument regarding the topic or no argument at all.
Label:
  \\\hline
\acs{iam} & What is the stance of the following sentence regarding the given topic? Only answer with one word; ``other'' if it is not a claim or unrelated to the topic, ``support'' only if it is a claim that supports the topic, or ``contest'' only if it is a claim that contests the topic. 
Sentence: [A sentence from the test set] 
Topic: [A respective topic from the test set] 
Label: 
& 
Decide the stance of the sentence below regarding the given topic. Unrelated sentences or non-claims are labelled with ``none'', supporting or contesting claims are labelled with ``support'' or ``contest''.
Sentence: [A sentence from the test set] 
Topic: [A respective topic from the test set] 
Label:
& 
Label the sentence ``[A sentence from the test set]'' with ``support'' if it is a claim that supports topic ``[A respective topic from the test set]'' or with ``contest'' if it is a claim that contests the topic. Label it "none" if the sentence is no claim regarding the topic or unrelated to it.
Label:
\\\hline
\acs{ibm} & Decide if the following sentence is a valid evidence with regard to the given claim. We define evidence as a sentence that clearly supports or contests the claim and is not merely a belief or a claim itself. Rather, an evidence provides an indication whether a claim is true. Only answer with one word; ``valid'' if the sentence is an evidence with regard to the claim, otherwise return ``invalid''.
Sentence: [A sentence from the test set] 
Claim: [A respective claim from the test set] 
Label:  
& 
Decide if the below sentence is a valid evidence regarding the given claim (label it ``yes'') or not (label it ``no''):
Sentence: [A sentence from the test set] 
Claim: [A respective claim from the test set] 
Label:
& 
Label the sentence ``[A sentence from the test set]'' with "valid" if it is a valid evidence for the claim ``[A respective claim from the test set]'' or "invalid" if not.
Label:
\\\hline
\Xhline{2\arrayrulewidth}
\end{tabularx}   
\caption{All prompts used for experiments with LLama2-70B and ChatGPT on \acs{fsc} (best: Prompt 2), \acs{iam} (best: Prompt 2), and \acs{ibm} (best: Prompt 1).}
\label{table_dataset_llm_phrases}
\end{table*}

\section{Dataset Licenses}\label{app:licenses}
We provide a list of all used datasets with their licenses:
\begin{itemize}[noitemsep]
\item \ac{fsc} (ours): Only annotations are included and licensed under CC-BY-SA 4.0. The annotated texts have to be extracted via script\footnotemark[2] from CommonCrawl (see also \ref{app:dataset_criteria}) or requested\footnotemark[1].
\item \ac{iam} \cite{cheng-etal-2022-iam}: The authors do not provide a license. The data is extracted from English Wikipedia.
\item \ac{ukpc} \cite{Stab2018b}: Licensed under CC-BY-NC.
\item \ac{ibm} \cite{ibm-corpuswide}: Licensed under CC-BY-SA 3.0.
\end{itemize}

\begin{sidewaystable*}\scriptsize
\centering

\begin{tabular}{lll | lll | lll}
\Xhline{2\arrayrulewidth}
 Topic & Most similar topic & Cosine sim. & Topic & Most similar topic & Cosine sim. & Topic & Most similar topic & Cosine sim. \\\hline

3d printer & holography & 0.11 & foreign aid & us intervention & 0.17 & prescription drug ads & big pharma & 0.19 \\
alcohol advertising & lower drinking age & 0.22 & fracking & offshore drilling & 0.21 & progressive tax & farm subsidies & 0.18 \\
alternative medicine & big pharma & 0.17 & free market & progressive tax & 0.16 & racial profiling & reverse discrimination & 0.18 \\
amazon & ebooks & 0.13 & freedom of speech & usa patriot act & 0.15 & religious holidays & atheism & 0.12 \\
anarchism & isolationism & 0.16 & fuel tax & progressive tax & 0.17 & renewable energy & wind energy & 0.20 \\
animal dissection & animal testing & 0.20 & gambling & legalized prostitution & 0.14 & reparations for slavery & white supremacy & 0.20 \\
animal testing & animal dissection & 0.20 & gay marriage & gay rights & 0.23 & reverse discrimination & white supremacy & 0.19 \\
antibiotic usage & alternative medicine & 0.16 & gay rights & gay marriage & 0.23 & right to health care & obamacare & 0.21 \\
artificial intelligence & autonomous cars & 0.12 & geothermal energy & hydroelectricity & 0.19 & robots & autonomous cars & 0.12 \\
assisted suicide & lethal injection & 0.22 & global warming & man-made greenhouse gases & 0.21 & sanctuary cities & illegal immigration & 0.20 \\
atheism & existence of god & 0.21 & glyphosate & gmos & 0.18 & school vouchers & charter schools & 0.24 \\
autonomous cars & lower speed limit & 0.19 & gmos & biofuels & 0.18 & sex education in school & birth control & 0.19 \\
beauty contest & feminism & 0.14 & government surveillance & usa patriot act & 0.20 & sex offender registry & mandatory sentencing & 0.18 \\
big pharma & prescription drug ads & 0.19 & guantanamo bay detention camp & drone strikes & 0.15 & smart home & smartwatch & 0.12 \\
bilingual education & standardized testing & 0.17 & holography & 3d printer & 0.11 & smartwatch & amazon & 0.12 \\
biofuels & offshore drilling & 0.19 & homeschooling & charter schools & 0.18 & social media & net neutrality & 0.13 \\
birth control & sex education in school & 0.19 & homework & homeschooling & 0.17 & solar energy & renewable energy & 0.19 \\
boarding school & charter schools & 0.17 & hydroelectricity & renewable energy & 0.20 & spanking & corporal punishment & 0.18 \\
border security & illegal immigration & 0.21 & illegal immigration & border security & 0.21 & sperm donor & surrogacy & 0.20 \\
brexit & foreign aid & 0.14 & insanity defense & mandatory sentencing & 0.20 & standardized testing & teacher tenure & 0.18 \\
bullfighting & factory farming & 0.13 & insider trading & labor unions & 0.13 & stem cell research & organ donation & 0.18 \\
cell phone radiation & stem cell research & 0.13 & isolationism & nuclear disarmament & 0.19 & surrogacy & sperm donor & 0.20 \\
censorship & net neutrality & 0.16 & jury duty & mandatory sentencing & 0.17 & suv & autonomous cars & 0.16 \\
charter schools & school vouchers & 0.24 & labor unions & unemployment insurance & 0.18 & teacher tenure & charter schools & 0.22 \\
cheerleading & beauty contest & 0.11 & legalized prostitution & monogamy & 0.18 & term limit & electoral college & 0.18 \\
clerical celibacy & monogamy & 0.18 & lethal injection & assisted suicide & 0.22 & tobacco advertising & electronic cigarettes & 0.20 \\
coal mining & fracking & 0.15 & libertarianism & right to health care & 0.16 & transgender rights & gay rights & 0.20 \\
community service & school vouchers & 0.12 & life extension & stem cell research & 0.14 & two-state solution & nuclear disarmament & 0.18 \\
compulsory voting & electoral college & 0.25 & lobbying & two-state solution & 0.14 & unemployment insurance & labor unions & 0.18 \\
concealed handguns & mandatory sentencing & 0.15 & lottery & crowdfunding & 0.13 & urban agriculture & farm subsidies & 0.17 \\
corporal punishment & mandatory sentencing & 0.19 & lower drinking age & alcohol advertising & 0.22 & urbanization & urban agriculture & 0.12 \\
crowdfunding & farm subsidies & 0.13 & lower speed limit & autonomous cars & 0.19 & us intervention & war on terrorism & 0.19 \\
cultured meat & factory farming & 0.20 & man-made greenhouse gases & global warming & 0.21 & usa patriot act & government surveillance & 0.20 \\
daycare & homeschooling & 0.14 & mandatory national service & right to health care & 0.15 & vaccination & animal testing & 0.15 \\
daylight saving time & solar energy & 0.11 & mandatory sentencing & insanity defense & 0.20 & vegetarianism & cultured meat & 0.20 \\
direct democracy & compulsory voting & 0.22 & monarchy & direct democracy & 0.16 & video games and violence & corporal punishment & 0.14 \\
drone strikes & war on terrorism & 0.20 & monogamy & gay marriage & 0.23 & virtual reality & artificial intelligence & 0.11 \\
ebooks & amazon & 0.13 & multiculturalism & white supremacy & 0.17 & voting machines & compulsory voting & 0.24 \\
ecotourism & urban agriculture & 0.16 & net neutrality & censorship & 0.16 & war on drugs & legalized prostitution & 0.17 \\
electoral college & compulsory voting & 0.25 & nuclear disarmament & isolationism & 0.19 & war on obesity & right to health care & 0.14 \\
electronic cigarettes & tobacco advertising & 0.20 & obamacare & right to health care & 0.21 & war on terrorism & drone strikes & 0.20 \\
executive order & obamacare & 0.18 & occupy wall street & white supremacy & 0.14 & water privatization & fracking & 0.15 \\
existence of god & atheism & 0.21 & offshore drilling & fracking & 0.21 & weather modification & man-made greenhouse gases & 0.16 \\
extraterrestrial life & existence of god & 0.14 & online dating service & monogamy & 0.10 & whaling & cultured meat & 0.16 \\
extreme sport & autonomous cars & 0.09 & organ donation & assisted suicide & 0.19 & white supremacy & reparations for slavery & 0.20 \\
factory farming & cultured meat & 0.20 & organic food & vegetarianism & 0.19 & wikileaks & government surveillance & 0.17 \\
farm subsidies & progressive tax & 0.18 & outsourcing & crowdfunding & 0.13 & wind energy & renewable energy & 0.20 \\
fast food & vegetarianism & 0.17 & pedelec & autonomous cars & 0.14 & wiretapping & government surveillance & 0.16 \\
felon voting & compulsory voting & 0.23 & plastic surgery & alternative medicine & 0.13 & women in the military & feminism & 0.18 \\
feminism & monogamy & 0.19 & police body cameras & government surveillance & 0.15 & year-round school & charter schools & 0.20 \\

       \Xhline{2\arrayrulewidth}
\end{tabular}

\caption{List of all 150 topics for the \ac{fsc}, including their semantically closest topic computed via embeddings with model ``all-MiniLM-L6-v2'' \cite{reimers-gurevych-2019-sentence}. Highest cosine similarity computed: 0.25.}\label{tbl:dataset_topics}
\end{sidewaystable*}

\begin{sidewaystable*}\scriptsize
\centering

\begin{tabular}{lll | lll | lll}
\Xhline{2\arrayrulewidth}
 \makecell[c]{Topic\\(\ac{fsc})} & \makecell[c]{Most similar topic\\(\ac{ukpc})} & Cosine sim. & \makecell[c]{Topic\\(\ac{fsc})} & \makecell[c]{Most similar topic\\(\ac{ukpc})} & Cosine sim. & \makecell[c]{Topic\\(\ac{fsc})} & \makecell[c]{Most similar topic\\(\ac{ukpc})} & Cosine sim. \\\hline

3d printer & minimum wage & 0.07 & fast food & minimum wage & 0.08 & online dating service & marijuana legalization & 0.04 \\
birth control & abortion & 0.17 & felon voting & death penalty & 0.15 & organ donation & cloning & 0.16 \\
alcohol advertising & marijuana legalization & 0.15 & feminism & abortion & 0.17 & organic food & marijuana legalization & 0.08 \\
alternative medicine & marijuana legalization & 0.12 & foreign aid & minimum wage & 0.12 & outsourcing & minimum wage & 0.12 \\
amazon & minimum wage & 0.05 & fracking & nuclear energy & 0.14 & pedelec & nuclear energy & 0.07 \\
anarchism & gun control & 0.11 & free market & minimum wage & 0.14 & plastic surgery & cloning & 0.10 \\
animal dissection & cloning & 0.12 & freedom of speech & gun control & 0.13 & police body cameras & gun control & 0.12 \\
animal testing & cloning & 0.15 & fuel tax & minimum wage & 0.12 & prescription drug ads & marijuana legalization & 0.14 \\
antibiotic usage & marijuana legalization & 0.09 & gambling & marijuana legalization & 0.12 & progressive tax & minimum wage & 0.16 \\
artificial intelligence & cloning & 0.10 & gay marriage & abortion & 0.14 & racial profiling & gun control & 0.13 \\
assisted suicide & death penalty & 0.17 & gay rights & abortion & 0.14 & religious holidays & school uniforms & 0.09 \\
atheism & cloning & 0.12 & geothermal energy & nuclear energy & 0.15 & renewable energy & nuclear energy & 0.15 \\
autonomous cars & gun control & 0.09 & global warming & nuclear energy & 0.14 & reparations for slavery & death penalty & 0.13 \\
beauty contest & school uniforms & 0.10 & glyphosate & marijuana legalization & 0.11 & reverse discrimination & school uniforms & 0.12 \\
big pharma & marijuana legalization & 0.14 & gmos & cloning & 0.14 & right to health care & abortion & 0.14 \\
bilingual education & school uniforms & 0.12 & government surveillance & gun control & 0.12 & robots & cloning & 0.09 \\
biofuels & nuclear energy & 0.14 & guantanamo bay detention camp & death penalty & 0.12 & sanctuary cities & marijuana legalization & 0.12 \\
birth control & abortion & 0.17 & concealed handguns & gun control & 0.20 & school vouchers & school uniforms & 0.15 \\
boarding school & school uniforms & 0.12 & holography & cloning & 0.05 & school vouchers & school uniforms & 0.15 \\
border security & marijuana legalization & 0.12 & homeschooling & school uniforms & 0.13 & sex education in school & abortion & 0.17 \\
brexit & minimum wage & 0.11 & homework & school uniforms & 0.13 & sex offender registry & death penalty & 0.14 \\
bullfighting & gun control & 0.08 & hydroelectricity & nuclear energy & 0.15 & smart home & nuclear energy & 0.07 \\
cell phone radiation & cloning & 0.10 & illegal immigration & marijuana legalization & 0.12 & smartwatch & minimum wage & 0.04 \\
censorship & gun control & 0.11 & insanity defense & death penalty & 0.17 & social media & school uniforms & 0.07 \\
charter schools & school uniforms & 0.15 & insider trading & minimum wage & 0.11 & solar energy & nuclear energy & 0.14 \\
cheerleading & school uniforms & 0.10 & isolationism & gun control & 0.12 & spanking & death penalty & 0.09 \\
clerical celibacy & abortion & 0.12 & jury duty & death penalty & 0.14 & sperm donor & cloning & 0.17 \\
stem cell research & cloning & 0.21 & labor unions & minimum wage & 0.16 & standardized testing & school uniforms & 0.14 \\
coal mining & nuclear energy & 0.11 & legalized prostitution & marijuana legalization & 0.15 & stem cell research & cloning & 0.21 \\
community service & minimum wage & 0.09 & lethal injection & death penalty & 0.22 & surrogacy & abortion & 0.17 \\
compulsory voting & minimum wage & 0.12 & libertarianism & abortion & 0.13 & suv & nuclear energy & 0.05 \\
concealed handguns & gun control & 0.20 & life extension & cloning & 0.12 & teacher tenure & school uniforms & 0.15 \\
corporal punishment & death penalty & 0.16 & lobbying & gun control & 0.13 & term limit & gun control & 0.12 \\
crowdfunding & minimum wage & 0.12 & lottery & minimum wage & 0.10 & tobacco advertising & marijuana legalization & 0.17 \\
cultured meat & cloning & 0.12 & lower drinking age & marijuana legalization & 0.15 & transgender rights & abortion & 0.13 \\
daycare & school uniforms & 0.08 & lower speed limit & gun control & 0.09 & two-state solution & gun control & 0.10 \\
daylight saving time & minimum wage & 0.09 & man-made greenhouse gases & nuclear energy & 0.13 & unemployment insurance & minimum wage & 0.18 \\
lethal injection & death penalty & 0.22 & mandatory national service & gun control & 0.12 & urban agriculture & minimum wage & 0.10 \\
direct democracy & gun control & 0.12 & mandatory sentencing & death penalty & 0.19 & urbanization & minimum wage & 0.10 \\
drone strikes & gun control & 0.14 & war on drugs & marijuana legalization & 0.19 & us intervention & gun control & 0.12 \\
ebooks & minimum wage & 0.05 & unemployment insurance & minimum wage & 0.18 & usa patriot act & gun control & 0.15 \\
ecotourism & nuclear energy & 0.10 & monarchy & abortion & 0.09 & vaccination & cloning & 0.12 \\
electoral college & gun control & 0.10 & monogamy & abortion & 0.13 & vegetarianism & abortion & 0.05 \\
electronic cigarettes & marijuana legalization & 0.13 & multiculturalism & school uniforms & 0.10 & video games and violence & gun control & 0.12 \\
executive order & abortion & 0.15 & net neutrality & minimum wage & 0.10 & virtual reality & cloning & 0.07 \\
existence of god & cloning & 0.12 & nuclear disarmament & nuclear energy & 0.18 & voting machines & gun control & 0.09 \\
extraterrestrial life & cloning & 0.12 & nuclear disarmament & nuclear energy & 0.18 & war on drugs & marijuana legalization & 0.19 \\
extreme sport & school uniforms & 0.07 & obamacare & minimum wage & 0.14 & war on obesity & marijuana legalization & 0.11 \\
factory farming & cloning & 0.12 & occupy wall street & gun control & 0.11 & war on terrorism & gun control & 0.13 \\
farm subsidies & minimum wage & 0.15 & offshore drilling & nuclear energy & 0.14 & water privatization & nuclear energy & 0.11 \\

       \Xhline{2\arrayrulewidth}
\end{tabular}

\caption{List of all 150 topics for the \ac{fsc} and their semantically closest topic from the \ac{ukpc}, computed via embeddings with model ``all-MiniLM-L6-v2'' \cite{reimers-gurevych-2019-sentence}. Highest cosine similarity computed: 0.22.}\label{tbl:dataset_topics_ukp_fs150t}
\end{sidewaystable*}

\section{Additional Figures}\label{app:sample_topic_figures_err}
\begin{figure*}
\centering
\begin{subfigure}{0.33\textwidth}
  \centering
  \includegraphics[width=1.0\linewidth]{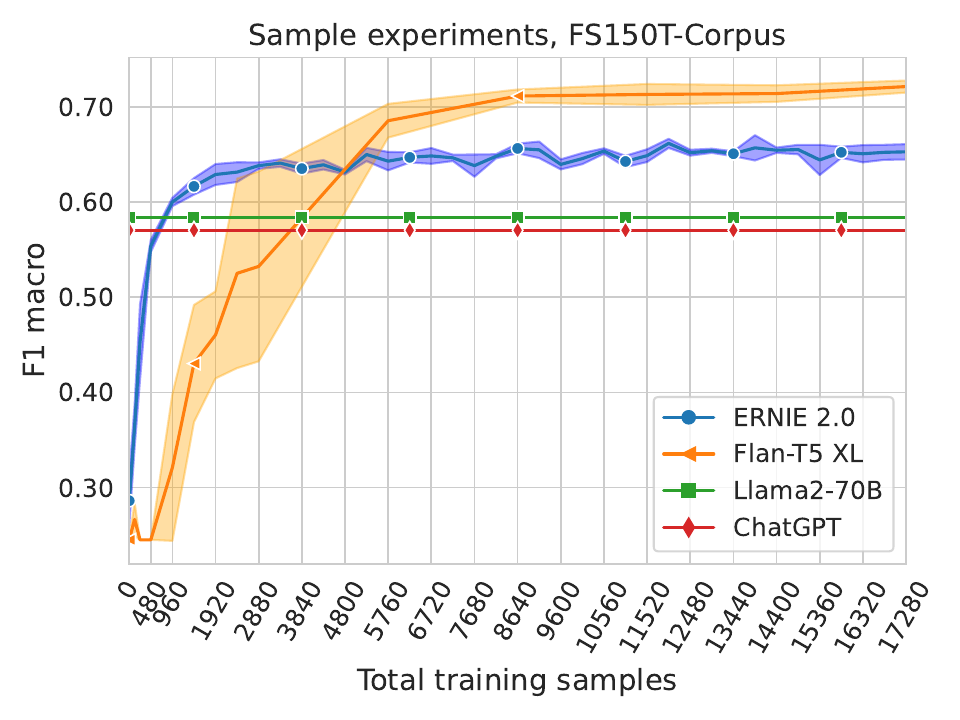}

  \label{fig:sample_experiment_plot_fsc_ernie}
\end{subfigure}%
\begin{subfigure}{0.33\textwidth}
  \centering
  \includegraphics[width=1.0\linewidth]{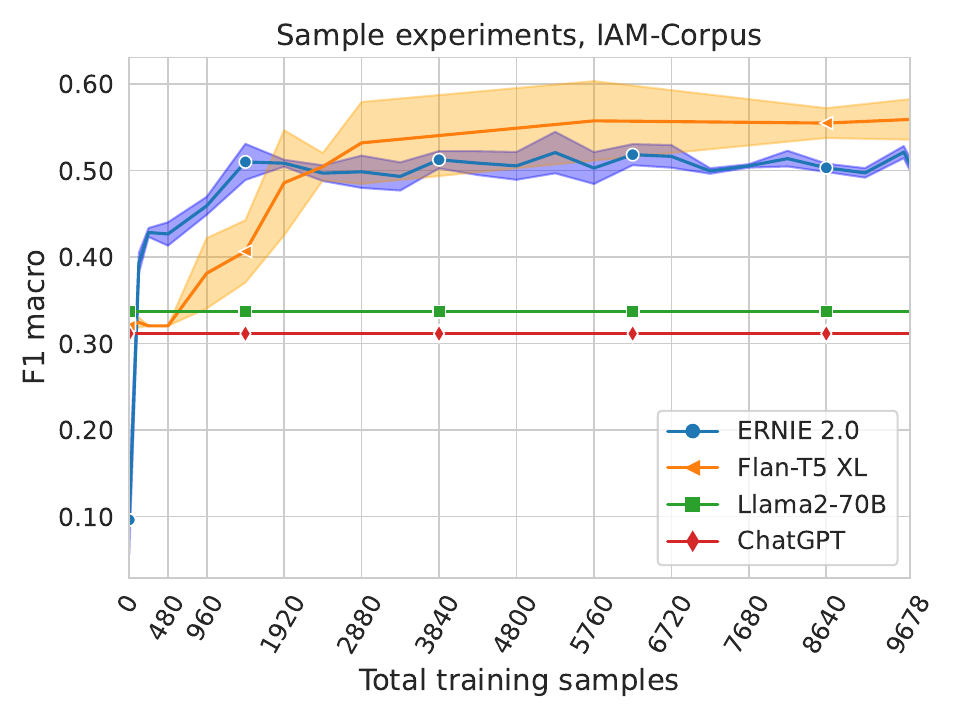}

  \label{fig:sample_experiment_plot_iam_ernie}
\end{subfigure}
\begin{subfigure}{0.33\textwidth}
  \centering
  \includegraphics[width=1.0\linewidth]{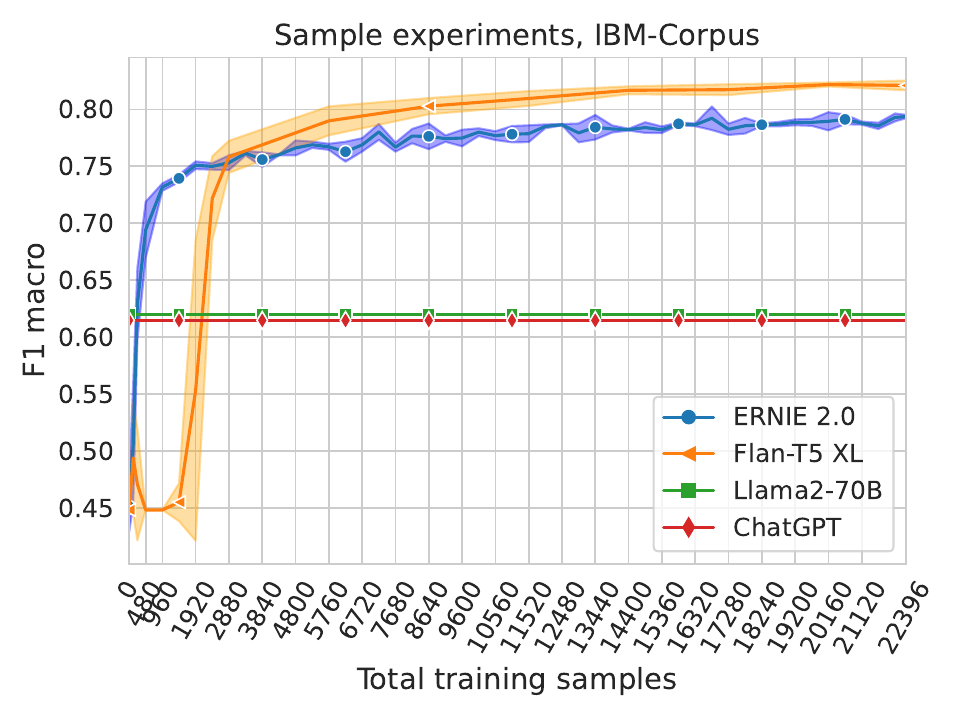}

  \label{fig:sample_experiment_plot_ibm_ernie}
\end{subfigure}

\caption{Sample experiments for FS150T-/IAM- and IBM-Corpus on ERNIE 2.0, FLAN-T5 XL, Llama2-70B, and ChatGPT in F${_1}$ macro and with standard deviation.}\label{fig:sample_experiment_plot_stddev}
\end{figure*}

\begin{figure*}
\centering
\begin{subfigure}{0.33\textwidth}
  \centering
  \includegraphics[width=1.0\linewidth]{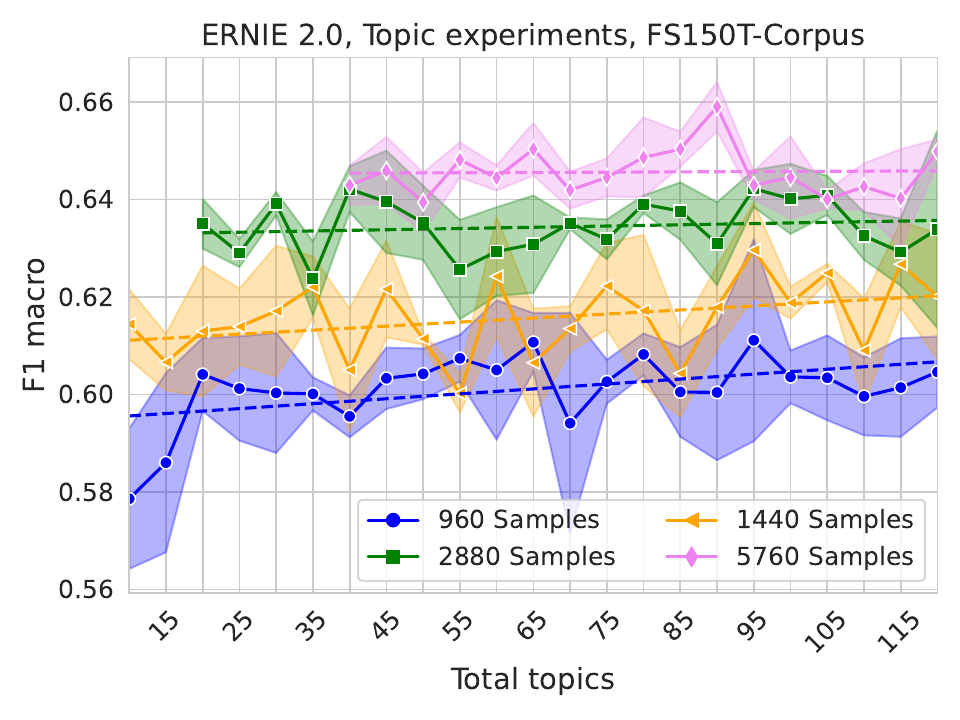}

  \label{fig:topic_experiment_plot_fsc_ernie_stddev}
\end{subfigure}%
\begin{subfigure}{0.33\textwidth}
  \centering
  \includegraphics[width=1.0\linewidth]{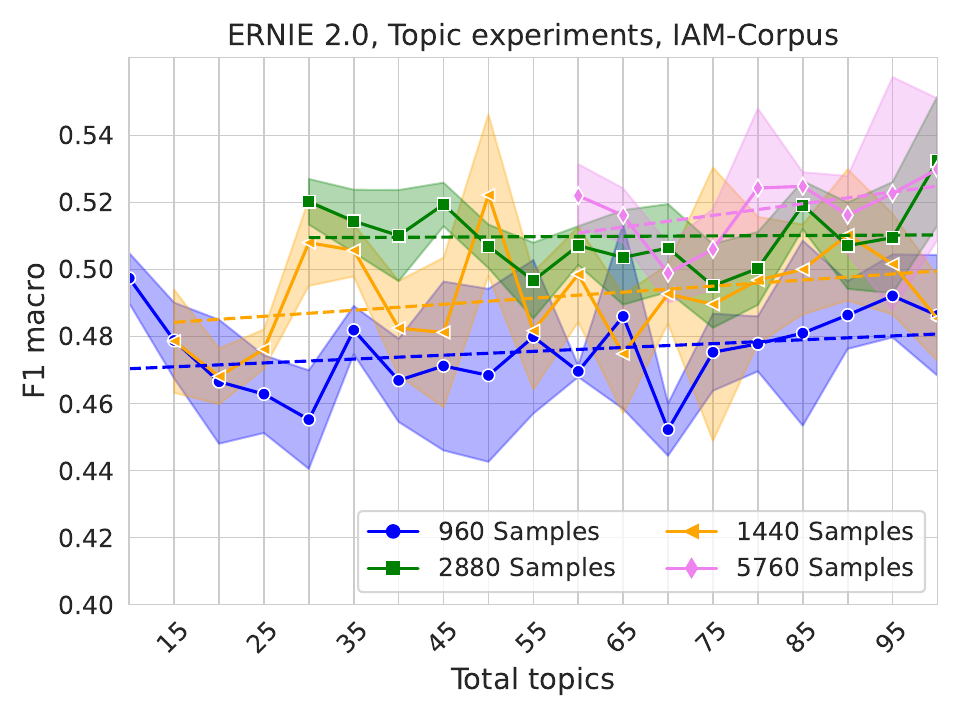}

  \label{fig:topic_experiment_plot_iam_ernie_stddev}
\end{subfigure}
\begin{subfigure}{0.33\textwidth}
  \centering
  \includegraphics[width=1.0\linewidth]{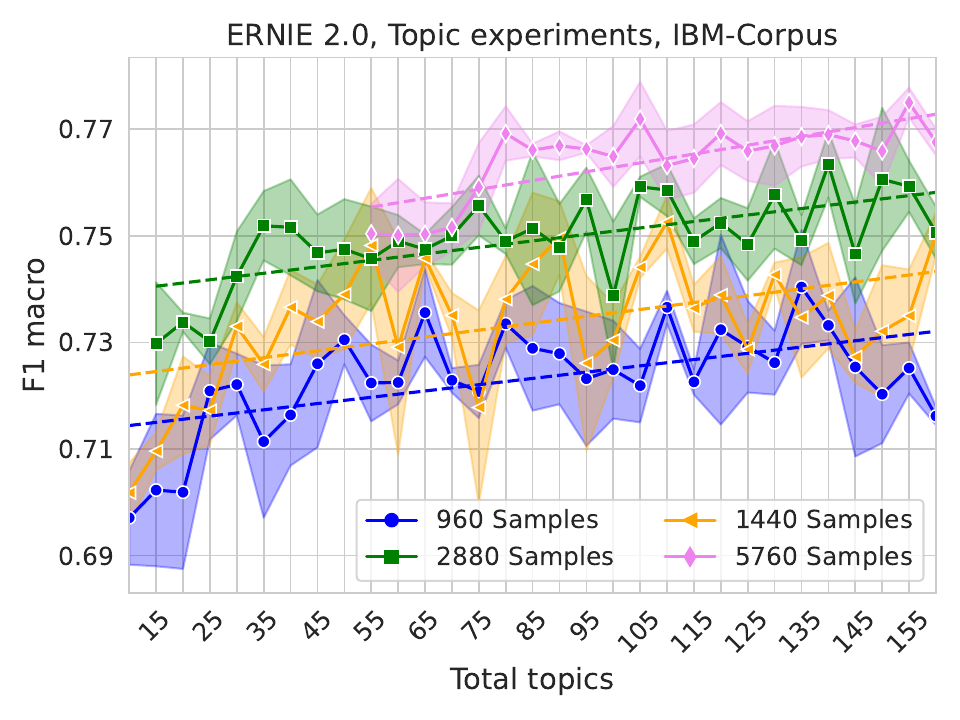}

  \label{fig:topic_experiment_plot_ibm_ernie_stddev}
\end{subfigure}

\begin{subfigure}{0.33\textwidth}
  \centering
  \includegraphics[width=1.0\linewidth]{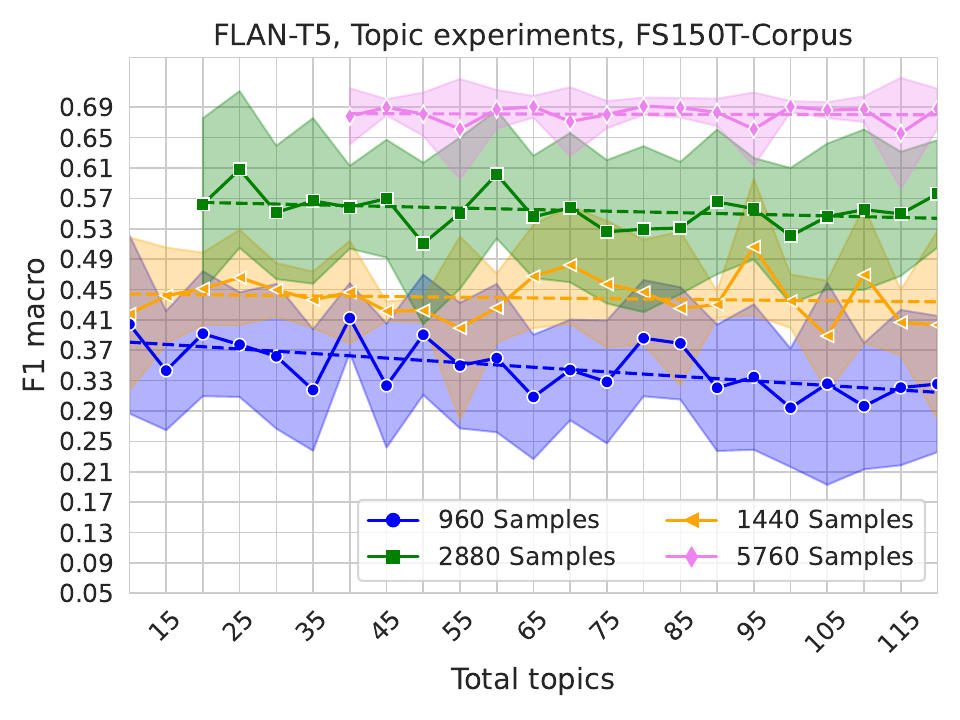}
  \label{fig:topic_experiment_plot_fsc_flan_stddev}
\end{subfigure}%
\begin{subfigure}{0.33\textwidth}
  \centering
  \includegraphics[width=1.0\linewidth]{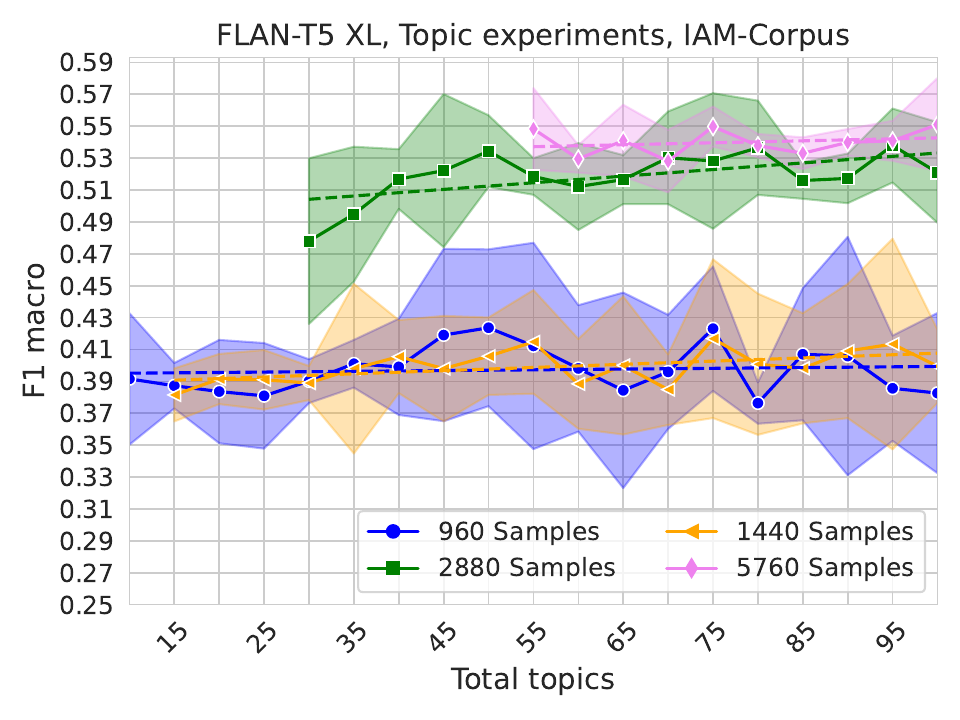}
  \label{fig:topic_experiment_plot_iam_flan_stddev}
\end{subfigure}
\begin{subfigure}{0.33\textwidth}
  \centering
  \includegraphics[width=1.0\linewidth]{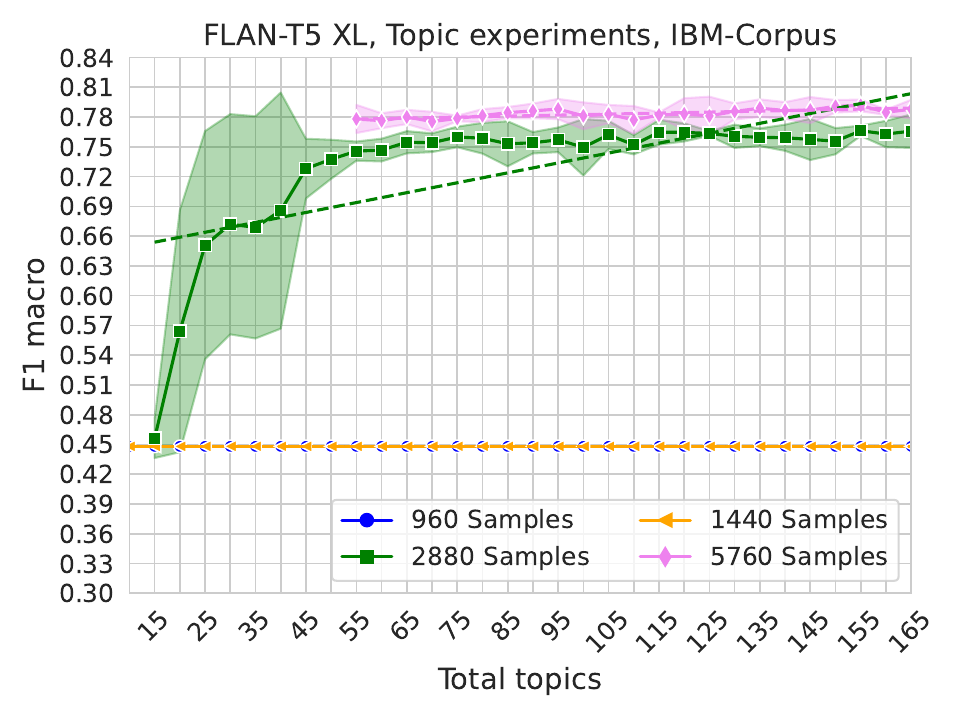}
  \label{fig:topic_experiment_plot_ibm_flan_stddev}
\end{subfigure}

\caption{Topic experiments for FS150T-/IAM- and IBM-Corpus on ERNIE 2.0 and FLAN-T5 XL in F${_1}$ macro and with standard deviation.}\label{fig:topic_experiment_plot_stddev}
\end{figure*}

This sections holds figures that include information about the standard deviation of sample (see Figures \ref{fig:sample_experiment_plot_stddev}) and topic experiments (see Figures \ref{fig:topic_experiment_plot_stddev}) that is left out in the main paper for better readability (see Sections \ref{sec:sample_exp} and \ref{sec:topic_exp}).

\end{document}